%% file: main.tex
\documentclass[10pt,twocolumn,letterpaper]{article}

\usepackage{cvpr}              

\usepackage[hyphens]{url}  
\usepackage{graphicx} 
\usepackage{natbib}  
\usepackage{caption} 
\usepackage{algorithm}
\usepackage{booktabs}
\usepackage{graphicx}
\usepackage{makecell}
\usepackage[table]{xcolor}
\usepackage{makecell}
\usepackage{amsmath}
\usepackage[table]{xcolor}
\usepackage{tabularx}
\usepackage{array}
\usepackage{adjustbox}
\usepackage{diagbox}
\usepackage{amsthm}

\definecolor{lightbluepurple}{HTML}{EEF0FA}
\usepackage{amsmath,amssymb}
\usepackage{algorithm}
\usepackage{algpseudocode}
\usepackage{xcolor}
\usepackage{colortbl}
\usepackage{booktabs}
\usepackage{array}
\usepackage{xcolor}
\usepackage{colortbl}
\usepackage{capt-of}
\usepackage{marvosym}

\definecolor{FWBlue}{HTML}{DCE9F8}

\usepackage{array}
\usepackage{tabularx}

\definecolor{FWLight}{HTML}{F5F8FC} 
\definecolor{FWBlue}{HTML}{C9DDF2}
\newcolumntype{Y}{>{\centering\arraybackslash}X}

\newcolumntype{M}{>{\hsize=2\hsize\linewidth=\hsize
  \raggedright\arraybackslash}X}
\newcolumntype{N}{>{\hsize=.75\hsize\linewidth=\hsize
  \centering\arraybackslash}X}
  
\definecolor{FWViolet}{HTML}{8070FF}

\colorlet{FWSRLight}{FWViolet!4!white}
\colorlet{FWSRDark}{FWViolet!22!white}
\usepackage{newfloat}
\usepackage{listings}
\DeclareCaptionStyle{ruled}{labelfont=normalfont,labelsep=colon,strut=off} 
\floatstyle{ruled}
\newfloat{listing}{tb}{lst}{}
\floatname{listing}{Listing}
\usepackage[table]{xcolor}
\usepackage{float}
\usepackage{placeins}

\input{preamble}
\definecolor{cvprblue}{rgb}{0.21,0.49,0.74}
\usepackage[pagebackref,breaklinks,colorlinks,allcolors=cvprblue]{hyperref}

\def\confName{CVPR}
\def\confYear{2027}

\title{Disentangling Visual Dynamics for World Model Planning}

\author{
Yuhong Shi$^{1}$ \quad
Yihan Guo$^{1}$ \quad
Jie Wei$^{2}$ \quad
Jianyi Liu$^{1}$ \Letter \quad
Jingwen Fu$^{3}$ \Letter \\[3pt]
{\small $^{1}$Institute of Artificial Intelligence and Robotics,
Xi'an Jiaotong University, Xi'an, China} \\
{\small $^{2}$China Mobile (Shanghai) Information and Communication
Technology Co., Ltd., Shanghai, China} \\
{\small $^{3}$Zhongguancun Academy, Beijing, China} \\[2pt]
{\small\ttfamily yuhongshi2023@stu.xjtu.edu.cn \quad
                 yeehanguo26@stu.xjtu.edu.cn \quad
                 weijie@sn.chinamobile.com } \\
{\small \Letter~\ttfamily jyliu@xjtu.edu.cn \quad
                 \Letter~fujingwen@bza.edu.cn}
}

\begin{document}
\maketitle
\input{sec/0_abstract}    
\input{sec/1_intro}

\input{sec/2_rela}

\input{sec/3_method}

\input{sec/4_experiment}
\input{sec/5_conclusion}


\section*{LLM Usage Statement}

LLMs assisted only in refining author-written text, code, and the presentation of author-developed ideas and conclusions. No reviewer-style manuscript feedback or other AI assistance was used. The authors take full responsibility for all submitted content.

{
    \small
    \bibliographystyle{ieeenat_fullname}
    \bibliography{main}
}
\input{sec/X_suppl}



\end{document}

%% file: sec/0_abstract.tex
\begin{abstract}


Visual world models enable planning through predictive latent states. However, entangled agent and exogenous dynamics can obscure task progress. Disentangling visual dynamics for planning requires addressing two challenges: 1) suppressing cross-factor interference while retaining useful information; and 2) linking factors to task-relevant signals, since stability alone does not ensure meaningful planning distances. Therefore, we introduce \textbf{FactorWorld}, a disentangled visual dynamics framework for world model planning. To separate agent and exogenous dynamics, we propose a Counterfactual Consensus Residual Factorizer (CCRF) guided by bidirectional counterfactual pairs. CCRF separates shared context from dynamic residuals, combining average and worst-partner constraints to suppress interference with state and motion supervision to retain useful information. To translate these factors into planning decisions, we introduce semantic routing to refine full-representation task predictions for state estimation, trajectory scoring, and task-adapted planning. Experiments on MiniGrid, MetaDrive, and CARLA show improved bidirectional factor stability across two backbones. On MiniGrid, factor routing raises mean SR from 78.80\% to 82.40\%. Across four continuous control tasks, including Push-T, FactorWorld achieves mean success rates ranging from 91.67\% to 100\% across tasks. These results demonstrate that combining decoupled visual dynamics with explicit tasks can significantly improve the planning capabilities of world models.

\end{abstract}

%% file: sec/1_intro.tex
\section{Introduction}
\label{sec:intro}

\begin{figure}[htbp]
  \centering
  \includegraphics[width=0.5\textwidth]{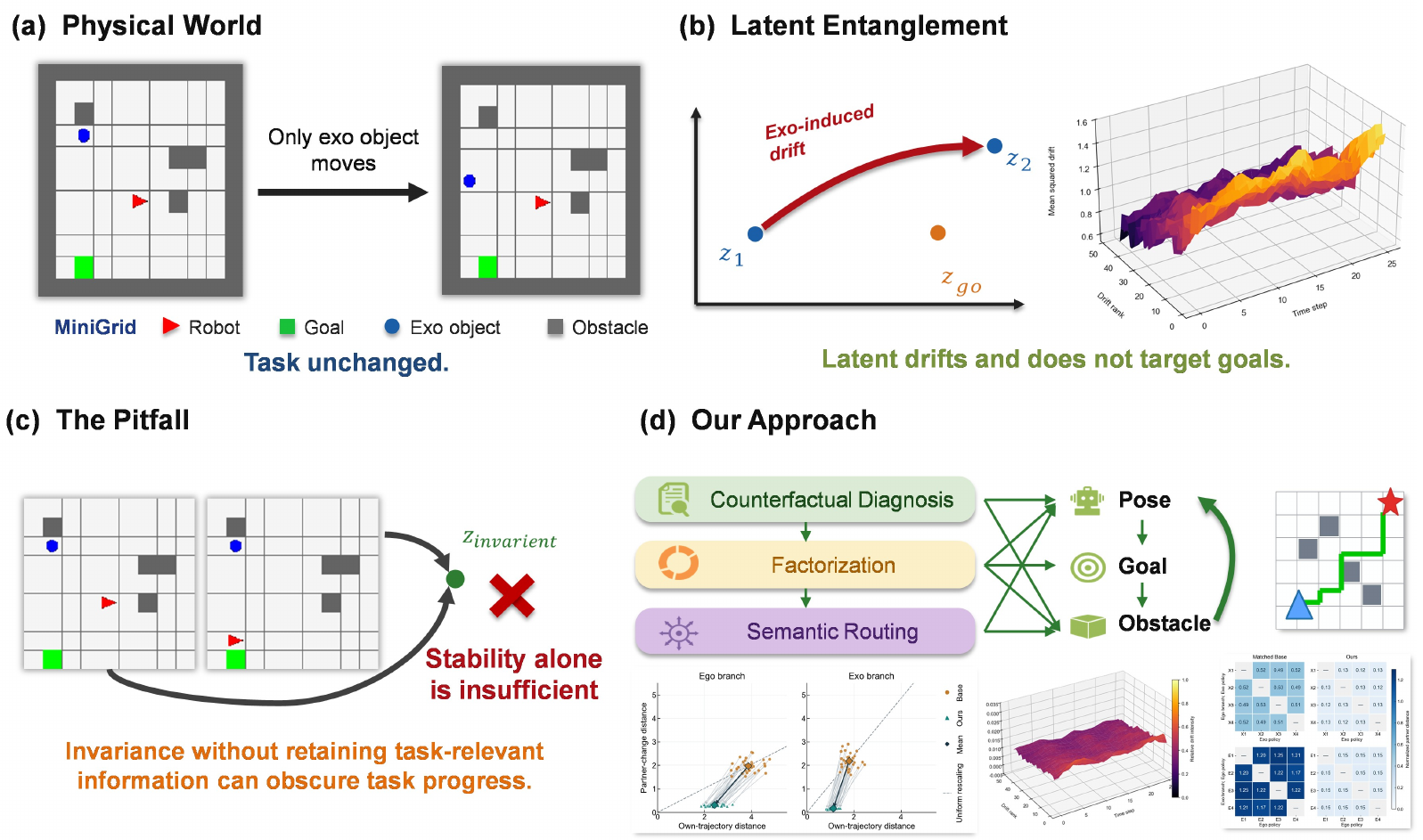}
  \caption{\textbf{From dynamic disentanglement to task-aware planning.}
(a--b) Exogenous motion induces latent drift unrelated to task progress.
(c) Invariance alone does not ensure planning utility.
(d) FactorWorld connects counterfactual disentanglement
with semantic routing for closed-loop planning.}
  \label{fig:motivation}
  \vspace{-1em}
\end{figure}

Visual world models encode complex observations into predictive
latent states that support planning~\cite{cheng2025jowa,achituve2026alewm}.
Yet representing environmental changes and organizing information
for decisions impose different requirements~\cite{Hutson2024Policy,Ma2024TransformerWorldModels}.
Visual changes may arise from agent motion, external object
motion, or both, with different implications for the
task~\cite{Hassan2025GEM}.
A representation rich in dynamic information may still fail to
distinguish task progress from changes that should be ignored
or explicitly considered~\cite{Hutson2024Policy}.
The key question is how to organize visual dynamics into
information that planning can selectively use.

This challenge becomes apparent in environments with exogenous
dynamics.
As illustrated in Fig.~\ref{fig:motivation}(a--b),
external motion can shift latent representations and their
distances to the goal even when the agent state and task goal
remain fixed, causing distance-based planning scores to
respond to changes unrelated to agent progress.
In LeWM~\cite{Maes2026LeWorldModel} experiments on
MiniGrid~\cite{Cheva2023MinigridMiniworld}, changing only
the exogenous object state produces measurable representation
drift.
This exposes a limitation of planning directly through latent
geometry, since the magnitude of representational change does
not inherently reflect an action's task value.
Addressing this mismatch requires distinguishing the sources
of dynamic variation and quantifying their interference.

Related approaches address this challenge through task-oriented
representation learning or explicit factor
separation~\cite{Wang2024MetaDT,wang2025disentangled}.
Task supervision encourages representations to retain information
useful for decisions~\cite{Hutson2024Policy,Roeder2025DALI},
whereas invariance and disentanglement objectives constrain
responses to nuisance variation~\cite{wang2025disentangled}.
These objectives address complementary properties, promoting
task information readability and stability under interference,
respectively.
Neither objective directly ensures that distances between
factor representations reflect goal reachability.
Fig.~\ref{fig:motivation}(c) illustrates why invariance
must be accompanied by information retention and an explicit
connection to the task. Therefore, we connect factor learning with its use in planning, preserving each factor's state and motion information and routing it into task predictions.

We introduce \textbf{FactorWorld}, a framework that connects
counterfactual dynamic disentanglement with semantic routing
for closed-loop visual planning.
Its Counterfactual Consensus Residual Factorizer (CCRF)
separates shared scene context from dynamic residuals and
learns Ego and Exo representations through average and
worst-partner stability constraints, complemented by state
and motion supervision.
Semantic routing uses these factors to refine task predictions
from the full representation, providing explicit interfaces
for state estimation, trajectory scoring, and task-adapted
planning.
We evaluate dynamic disentanglement and closed-loop planning
in MiniGrid~\cite{Cheva2023MinigridMiniworld},
MetaDrive~\cite{li2023metadrive}, and
CARLA~\cite{dosovitskiy2017carla}, and further evaluate
continuous control adaptations on Two-Room~\cite{sobal2025learning}, Reacher~\cite{tassa2018deepmindcontrolsuite},
Push-T~\cite{florence2021implicit}, and Cube~\cite{ogbench_park2025}.
On MiniGrid, factor routing raises mean planning SR
from 78.80\% to 82.40\%; the continuous control adaptations
achieve mean SRs ranging from 91.67\% to 100\%.
Our main contributions are summarized as follows:
\begin{itemize}
    \item {Counterfactual diagnosis of dynamic entanglement.}
    We use bidirectional trajectory pairing to quantify
    cross-factor interference, distinguishing responses
    to each factor's own variation from sensitivity
    to its partner.
    
    \item {Dynamic disentanglement with information retention.}
    We introduce CCRF with average and worst-partner constraints, complemented by state and motion supervision, to learn selective
    dynamic representations.
    
    \item {Semantic routing for visual planning.}
    We use factor representations to refine task predictions
    from the full representation, connecting dynamic
    disentanglement to explicit decision interfaces for
    visual navigation and task-adapted continuous control.
\end{itemize}

%% file: sec/2_rela.tex
\section{Related Work}
\label{sec:Related}



\subsection{Latent World Models for Visual Planning}

Latent world models connect visual representations to planning by predicting future states under candidate actions~\cite{huang2026pointworld,Bi2026Motus,kim2026compact}. 
DINO-WM~\cite{zhou2025dinowm} predicts pretrained visual features, while LeWorldModel~\cite{Maes2026LeWorldModel} jointly learns an encoder and latent dynamics from pixels. Both score action sequences by matching predicted terminal representations to a goal embedding. 
Other methods use task-specific interfaces to connect predictions to decisions. 
Drive-WM~\cite{wang2024driving} evaluates generated multi-view futures using image-based rewards, while Navigation World Models~\cite{Bar_2025_CVPR} simulates outcomes to construct or rank trajectories. World4Drive~\cite{zheng2025world4drive} evaluates candidates through latent futures conditioned on driving intentions, and DriveLaW~\cite{xia2026drivelaw} feeds video generation latents into a learned planner. 
However, predictable latent changes do not necessarily indicate task progress. When agent and exogenous dynamics are mixed, external motion can alter goal distance scores without changing the agent's pose or goal. This motivates us to study how different dynamic factors contribute to planning scores and task readouts. 

\subsection{Disentangled Representations for Control}

Disentangled representations support visual control by separating agent dynamics from exogenous variation while retaining information for decisions \cite{yao2026towards,tang2026causalvad}. 
DisWM~\cite{wang2025disentangled} transfers knowledge from distracting videos through latent distillation and disentanglement constraints. Counterfactual learning provides complementary supervision. SC-CF~\cite{kung2025changed} uses hypothetical state changes for video representation learning, while CIDiffuser~\cite{huang2025visual} applies causal intervention to controllable image editing. E-CWM~\cite{yao2026towards} combines observation perturbations and counterfactual substitutions for navigation adaptation, and RDCL~\cite{qi2025robust} separates static and dynamic factors for physical reasoning. 
These methods use factor separation and counterfactual supervision to improve transfer, adaptation, or reasoning, but do not directly constrain interference between agent and exogenous trajectories or specify how the separated factors should affect planning scores \cite{Bi2026Motus}. Invariance alone can also discard information needed for control. 
We therefore introduce CCRF to suppress average and worst-partner interference while preserving each factor's state and motion, and semantic routing to translate these factors into pose, goal, and occupancy predictions for planning.
 

%% file: sec/3_method.tex
\begin{figure*}[t]
  \centering
  \includegraphics[width=\textwidth]{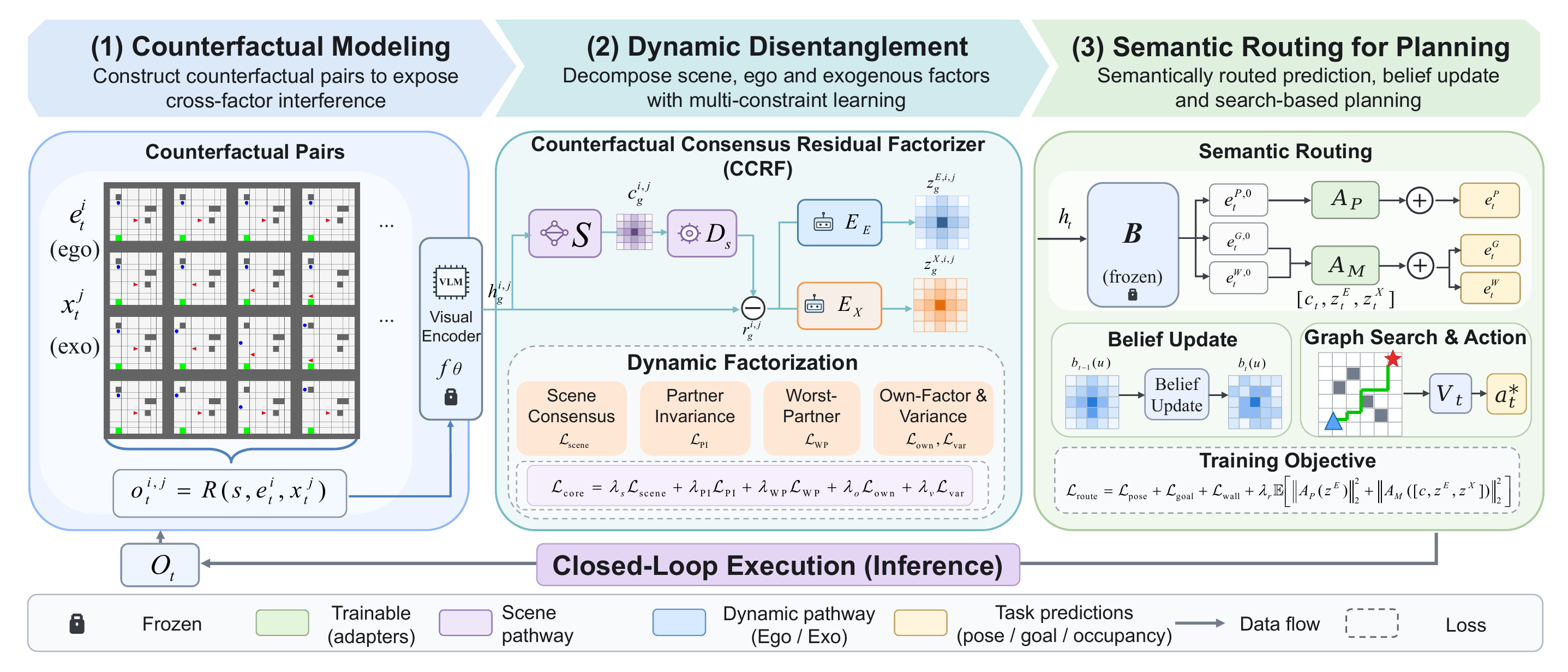}
  \caption{\textbf{Overview of the FactorWorld framework.}
  Counterfactual pairs expose interference between agent and
  exogenous motion. CCRF separates scene context and dynamic
  factors, which refine pose, goal, and occupancy predictions
  through semantic routing. These predictions support belief
  updates and graph search for closed-loop planning.}
  \label{fig:framework}
  \vspace{-1em}
\end{figure*}

\section{Method}


\subsection{Counterfactual Modeling}
\label{sec:counterfactual_modeling}

We consider visual goal-directed navigation with exogenous dynamics. We model an observation as $o_t=\mathcal{R}(s,e_t,x_t)$, where $s$, $e_t$, and $x_t$ denote the static scene (scene), agent state (ego), and external object state (exo), respectively. 
A visual encoder $f_\theta$ and a disentanglement module
$F_\phi$, parameterized by $\theta$ and $\phi$, extract
\begin{equation}
    h_t=f_\theta(o_t), \qquad
    (c_t,z_t^E,z_t^X)=F_\phi(h_t).
\end{equation}
where $h_t$ is the full visual representation, and
$c_t$, $z_t^E$, and $z_t^X$ represent scene context,
agent dynamics, and exogenous dynamics, respectively.
Our objective is to preserve each dynamic factor's variation while remaining invariant to changes in the other factor, thereby reducing interference from exogenous motion in task scoring.

To identify cross-factor interference, we cross $K$ agent trajectories with $L$ exogenous trajectories within the same scene, yielding time-aligned counterfactual observations
$o_g^{i,j}=\mathcal{R}(s,e_t^i,x_t^j)$.
Here $g=(s,t)$ indexes a scene--time group,
$i\in\{1,\ldots,K\}$ and $j\in\{1,\ldots,L\}$ index agent
and exogenous trajectories, and $e_t^i$ and $x_t^j$ denote
their respective states at time $t$.
Holding $i$ fixed while varying $j$ preserves the agent state
and changes the exogenous state. 
The reverse pairing preserves the exogenous state.
We define dynamic entanglement as a factor's representation
changing in response to another factor while its own state
remains fixed. These pairs distinguish own-factor sensitivity from partner
interference to guide factor learning.

\subsection{Dynamic Disentanglement}
\label{sec:dynamic_factorization}


CCRF learns selective dynamic representations that remain
stable under partner changes while retaining information
about their own state and motion.
Given a frozen visual representation
$h_g^{i,j}=f_\theta(o_g^{i,j})$, a scene encoder $S$ extracts
shared context, and a mapping $D_s$ projects it back to the
full latent space to form a dynamic residual
\begin{equation}
\label{eq:scene_residual}
    c_g^{i,j}=S(h_g^{i,j}), \qquad
    r_g^{i,j}=h_g^{i,j}-D_s(c_g^{i,j}).
\end{equation}
Separate encoders $E_E$ and $E_X$ extract Ego and Exo factors
\begin{equation}
\label{eq:dynamic_branches}
    z_g^{E,i,j}=E_E(r_g^{i,j}), \qquad
    z_g^{X,i,j}=E_X(r_g^{i,j}).
\end{equation}
The scene consensus loss $\mathcal{L}_{\mathrm{scene}}$
aligns scene codes with their group mean
$\bar{c}_g=(KL)^{-1}\sum_{i,j}c_g^{i,j}$.
This structure separates shared context from residual
variation, while counterfactual constraints establish
the roles of the two dynamic branches.

For branch $q\in\{E,X\}$, let $a$ and $b$ index its own
and partner trajectories, respectively.
We reindex $(a,b)=(i,j)$ for Ego and $(a,b)=(j,i)$ for Exo.
Define the conditional mean and partner-induced deviation as
\begin{equation}
\label{eq:conditional_deviation}
    \bar{z}_{g,a}^{q}
    =\frac{1}{B_q}\sum_b z_{g,a,b}^{q}, \qquad
    \delta_{g,a,b}^{q}
    =z_{g,a,b}^{q}-\bar{z}_{g,a}^{q},
\end{equation}
where $B_E=L$ and $B_X=K$ are the numbers of partners.
We jointly constrain average interference and large
partner deviations through
\begin{equation}
\label{eq:partner_invariance}
    \mathcal{L}_{\mathrm{PI}}
    =
    \sum_{q\in\{E,X\}}
    \mathbb{E}_{g,a,b}
    \left[\frac{\|\delta_{g,a,b}^{q}\|_2^2}{d_q}\right],
\end{equation}
\begin{equation}
\label{eq:worst_partner}
    \mathcal{L}_{\mathrm{WP}}
    =
    \sum_{q\in\{E,X\}}
    \mathbb{E}_{g,a}
    \left[
        \tau\log\sum_b
        \exp\!\left(\frac{\rho_{g,a,b}^{q}}{\tau}\right)
    \right],
\end{equation}
where $d_q$ is the branch dimension and
$\rho_{g,a,b}^{q}
=\min\{M,\sqrt{\|\delta_{g,a,b}^{q}\|_2^2/d_q+\epsilon}\}$
is a stabilized RMS deviation with $\epsilon>0$ and cap $M>0$.
Expectations are uniformly averaged over the indexed groups
and trajectories.
The temperature $\tau>0$ controls the emphasis on larger
deviations, complementing average invariance with stronger
constraints on disruptive partners.

To retain useful information, $\mathcal{L}_{\mathrm{own}}$
supervises state and displacement readouts from individual
factors and their conditional means, together with action
prediction from the Ego conditional mean.
A variance regularizer $\mathcal{L}_{\mathrm{var}}$
discourages collapse.
The core objective is
\begin{equation}
\label{eq:factorization_core}
\begin{aligned}
    \mathcal{L}_{\mathrm{core}}
    ={}&
    \lambda_s\mathcal{L}_{\mathrm{scene}}
    +\lambda_{\mathrm{PI}}\mathcal{L}_{\mathrm{PI}}
    +\lambda_{\mathrm{WP}}\mathcal{L}_{\mathrm{WP}} \\
    &+\lambda_o\mathcal{L}_{\mathrm{own}}
    +\lambda_v\mathcal{L}_{\mathrm{var}},
\end{aligned}
\end{equation}
where the nonnegative $\lambda$ coefficients balance
the corresponding losses.
These objectives jointly suppress partner interference
and encourage retention of state and motion information.
Counterfactual pairs and supervision labels are used only
during training; inference extracts all factors from a
single observation for downstream planning.


\subsection{Semantic Routing for Planning}
\label{sec:semantic_routing}

Factor stability does not ensure that latent distances reflect
goal reachability.
We therefore retain a base task readout from the full
representation and introduce factor corrections through
semantically restricted residual pathways.
We describe the navigation implementation below;
task-specific control adaptations are detailed in the appendix.

A frozen readout $B$ maps $h_t$ to base logits
$(\ell_t^{P,0},\ell_t^{G,0},\ell_t^{W,0})$
for agent pose, goal location, and obstacle occupancy.
Residual adapters $A_P$ and $A_M$ refine these predictions
\begin{equation}
\label{eq:semantic_routing}
\begin{aligned}
    \ell_t^P
    &= \ell_t^{P,0}+A_P(z_t^E), \\
    [\ell_t^G,\ell_t^W]
    &= [\ell_t^{G,0},\ell_t^{W,0}]
       +A_M([c_t,z_t^E,z_t^X]),
\end{aligned}
\end{equation}
where brackets denote concatenation.
The Ego factor corrects position and orientation, while
the combined factors correct goal and occupancy predictions.
This routing preserves the full representation's base
predictions while assigning factor corrections to specific
task outputs.
The adapters' final layers are initialized to zero.
With the encoder, factorizer, and base readout frozen,
we train only the adapters using
\begin{equation}
{\small
\label{eq:routing_loss}
\begin{aligned}
    \mathcal{L}_{\mathrm{route}}
    ={}& \mathcal{L}_{\mathrm{pose}}
        +\mathcal{L}_{\mathrm{goal}}
        +\mathcal{L}_{\mathrm{wall}} \\
       &+\lambda_r\mathbb{E}\!\left[
         \|A_P(z^E)\|_2^2
         +\|A_M([c,z^E,z^X])\|_2^2
       \right],
\end{aligned}
}
\end{equation}
where, $\mathcal{L}_{\mathrm{pose}}$ and
$\mathcal{L}_{\mathrm{goal}}$ are classification losses,
$\mathcal{L}_{\mathrm{wall}}$ is a binary occupancy loss,
and $\lambda_r\geq0$ controls residual regularization.
The expectation averages over training observations;
spatial labels are used only during training.

To integrate predictions over time, we maintain a belief
$b_t(u)$ over poses $u=(p_x,p_y,\theta)$, comprising position
and orientation.
The previous action propagates the belief, and the pose
readout supplies the current observation score
\begin{equation}
\label{eq:belief_update}
\begin{aligned}
    b_t^{-}(u)
    &= \sum_{u'}
       T_{\bar{W}_{t-1}}(u\mid u',a_{t-1})b_{t-1}(u'), \\
    b_t(u)
    &\propto
       \big[(1-\varepsilon)b_t^{-}(u)+\varepsilon U(u)\big]
       L_t(u)^\alpha,
\end{aligned}
\end{equation}
where, $b_t^{-}$ is the predicted belief and
$T_{\bar{W}_{t-1}}$ combines known action rules with the
accumulated occupancy estimate $\bar{W}_{t-1}$.
$L_t(u)$ is the pose observation score, $U$ is uniform,
$\varepsilon\in[0,1]$ provides smoothing, and $\alpha>0$
controls observation weighting.
We accumulate occupancy predictions by exponential averaging
and estimate the goal from the goal image and observed history.

The predicted occupancy defines a pose graph with turn
and forward-motion edges, including obstacle penalties.
Dijkstra search computes the remaining cost $V_t$ to
the estimated goal $\hat{g}_t$.
From the most probable pose
$\hat{u}_t=\arg\max_u b_t(u)$, we select
\begin{equation}
{\small
\label{eq:routed_action}
\begin{aligned}
    a_t^*=\arg\min_a \Big[
        &\kappa_t(\hat{u}_t,a)
        +V_t\!\left(f(\hat{u}_t,a)\right) \\
        &+\lambda_{\mathrm{visit}}
          \max\!\left(N_t(f(\hat{u}_t,a))-1,0\right)
    \Big],
\end{aligned}}
\end{equation}
where, $f$ is the discrete pose transition, $\kappa_t$
is the immediate action cost, and $N_t$ counts visits
to estimated poses.
The weight $\lambda_{\mathrm{visit}}\geq0$ penalizes
repeated visits to discourage cycles.
After executing one action, we update the belief and
predicted graph from the new observation and replan.
Thus, factor information enters closed-loop decisions
through explicit task predictions rather than latent
distance alone.

%% file: sec/4_experiment.tex
\begin{figure*}[htbp]
  \centering
  \includegraphics[width=\textwidth]{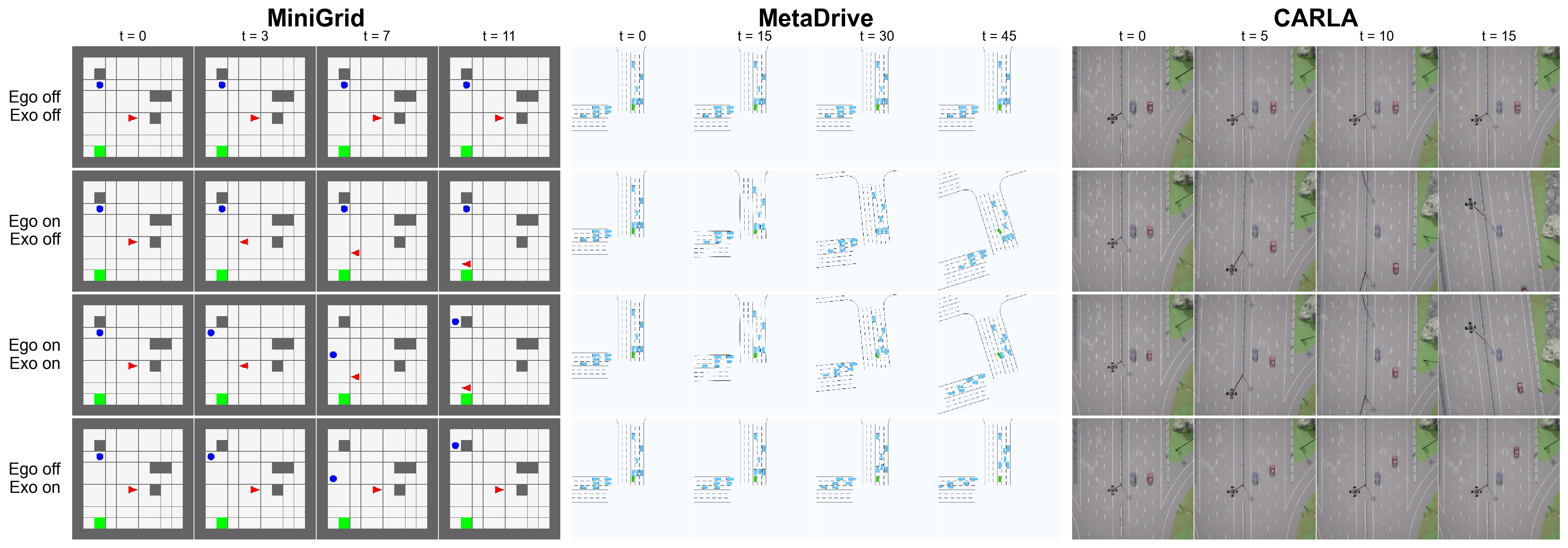}
\caption{\textbf{Controlled dynamic variations across three environments.}
Rows show four combinations of agent (Ego) and external object
(Exo) motion, with each enabled or disabled independently.
Columns within each environment show successive observations.
}\label{fig:counterfactual_examples}
  \label{fig:p3}
\end{figure*}

\begin{table*}[htbp]
\centering
\begin{minipage}[t]{0.68\textwidth}
\vspace{0pt}
\caption{\textbf{Counterfactual diagnosis of dynamic entanglement.}
$E_A$ and $E_B$ denote Ego and Exo main-effect energies.
Reduction is relative to the corresponding backbone baseline.
Bold marks the lowest ratio and highest reduction within
each backbone and environment.
State supervision is added to Ego projection.}
\label{tab:entanglement_diagnosis}
\footnotesize
\setlength{\tabcolsep}{3pt}
\renewcommand{\arraystretch}{1.12}
\begin{tabularx}{\linewidth}{
  @{}l
  *{4}{>{\centering\arraybackslash}X}
  @{}
}
\toprule
\textbf{Method}
& \textbf{Ego} $E_A$
& \textbf{Exo} $E_B$
& $E_B/E_A$ $\downarrow$
& \textbf{Reduction} $\uparrow$ \\
\midrule
\multicolumn{5}{c}{\textbf{\textit{(a) MiniGrid}}} \\
\midrule
LeWM \cite{Maes2026LeWorldModel}
& 164.7907 & 62.1676 & 0.3773 & -- \\
\quad + Ego proj.
& 51.6443 & 1.1574 & 0.0224 & 94.06\% \\
\quad + Ego proj. + State sup.
& 55.8291 & 0.6161 & \textbf{0.0110} & \textbf{97.07\%} \\
JEPA \cite{assran2023ijepa}
& 0.0597 & 0.0252 & 0.4217 & -- \\
\quad + Ego proj.
& 0.0350 & 0.0135 & 0.3844 & 8.83\% \\
\quad + Ego proj. + State sup.
& 0.0365 & 0.0140 & \textbf{0.3820} & \textbf{9.42\%} \\
\midrule
\multicolumn{5}{c}{\textbf{\textit{(b) MetaDrive}}} \\
\midrule
LeWM \cite{Maes2026LeWorldModel}
& 372.5075 & 183.4601 & 0.4925 & -- \\
\quad + Ego proj.
& 77.0057 & 0.2617 & \textbf{0.0034} & \textbf{99.31\%} \\
\quad + Ego proj. + State sup.
& 69.5014 & 0.3259 & 0.0047 & 99.05\% \\
JEPA \cite{assran2023ijepa}
& 372.5075 & 183.4601 & 0.4925 & -- \\
\quad + Ego proj.
& 106.6947 & 0.7535 & \textbf{0.0071} & \textbf{98.57\%} \\
\quad + Ego proj. + State sup.
& 89.3080 & 1.0374 & 0.0116 & 97.64\% \\
\midrule
\multicolumn{5}{c}{\textbf{\textit{(c) CARLA}}} \\
\midrule
LeWM \cite{Maes2026LeWorldModel}
& 371.8022 & 34.2827 & 0.0922 & -- \\
\quad + Ego proj.
& 65.1661 & 0.7739 & \textbf{0.0119} & \textbf{87.12\%} \\
\quad + Ego proj. + State sup.
& 46.4477 & 1.1965 & 0.0258 & 72.06\% \\
JEPA \cite{assran2023ijepa}
& 206.5511 & 12.8489 & 0.0622 & -- \\
\quad + Ego proj.
& 74.5412 & 0.6043 & 0.0081 & 86.97\% \\
\quad + Ego proj. + State sup.
& 82.7798 & 0.3094 & \textbf{0.0037} & \textbf{93.99\%} \\
\bottomrule
\end{tabularx}
\end{minipage}
\hfill
\begin{minipage}[t]{0.30\textwidth}
\vspace{0pt}
\centering
\includegraphics[
  width=1\linewidth,
  height=\textheight,
  keepaspectratio
]{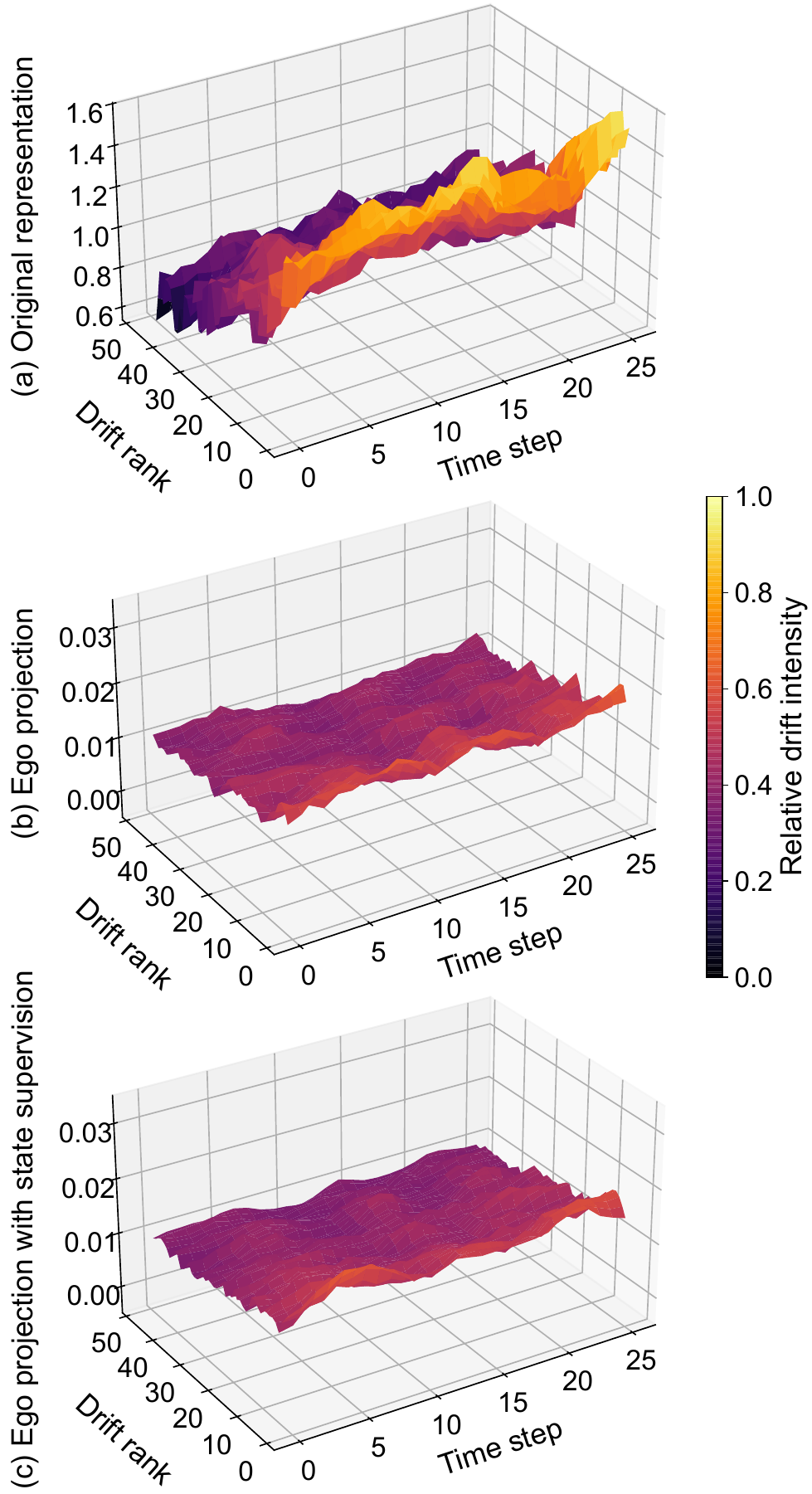}
\captionof{figure}{\textbf{Exogenous latent drift.}
Mean squared differences under exogenous changes with
matched agent states. Vertical scales differ across panels.}
\label{fig:latent_drift}
\end{minipage}
\end{table*}

\section{Experiment}

\subsection{Experimental Setup}



\textbf{Datasets.}
We use MiniGrid \cite{Cheva2023MinigridMiniworld} with moving exogenous objects as the primary environment for closed-loop planning, and evaluate factorization across environments using MetaDrive \cite{li2023metadrive} and CARLA \cite{dosovitskiy2017carla}. Counterfactual data cross four agent trajectories with four exogenous trajectories
within each fixed scene, yielding $4\times4$ branches.
The MiniGrid factorization dataset contains 320 maps, split into 224/32/64 for training, validation, and testing. MetaDrive contains 64 scene groups with a 48/16 training/validation
split, while CARLA contains 64 groups with a 48/8/8 training/validation/test split. For each environment, we freeze the corresponding LeWM encoder and train the factorizer separately.
Fig.~\ref{fig:counterfactual_examples} illustrates observation
sequences with agent and exogenous motion independently
enabled or disabled across the three environments.
These controlled variations distinguish changes caused by
each dynamic factor.

\textbf{Metrics.}
We quantify dynamic entanglement using NOR and Group-IRS.
For either dynamic branch, $z_{g,a,b}$ denotes its
representation in scene--time group $g=(s,t)$,
where $a$ indexes its own trajectory and $b$ its partner.
For Ego, $(a,b)=(i,j)$, for Exo, $(a,b)=(j,i)$.
NOR measures partner-induced variation relative to
own-factor variation
\begin{equation}
\label{eq:nor}
    \operatorname{NOR}
    =
    \frac{
        \mathbb{E}_{g,a,b<b'}
        \|z_{g,a,b}-z_{g,a,b'}\|_2
    }{
        \mathbb{E}_{g,b,a<a'}
        \|z_{g,a,b}-z_{g,a',b}\|_2
    },
\end{equation}
were, $\|\cdot\|_2$ is the Euclidean norm, and each
expectation uniformly averages over the indexed groups
and trajectory pairs.
The restrictions $b<b'$ and $a<a'$ count each distinct
unordered pair once.
A lower NOR indicates weaker relative partner interference.

Let $A$ and $B$ be the numbers of own and partner trajectories,
with $(A,B)=(K,L)$ for Ego and $(A,B)=(L,K)$ for Exo.
Define the conditional mean
$\mu_{g,a}=B^{-1}\sum_b z_{g,a,b}$
and group mean
$\mu_g=(AB)^{-1}\sum_{a,b}z_{g,a,b}$.
Group-IRS measures worst-partner stability
{\footnotesize
\begin{equation}
\label{eq:group_irs}
    \operatorname{GroupIRS}
    =
    \mathbb{E}_g\!\left[
        1-
        \frac{
            A^{-1}\sum_a\max_b
            \|z_{g,a,b}-\mu_{g,a}\|_2
        }{
            \max_{a,b}\|z_{g,a,b}-\mu_g\|_2
        }
    \right].
\end{equation}
}
The numerator averages the largest partner-induced deviation
for each own trajectory, while the denominator measures
the maximum distance from the group mean.
$\mathbb{E}_g$ averages uniformly over groups.
Higher Group-IRS indicates smaller worst-partner deviations
relative to the within-group representation spread.
Both metrics are evaluated separately for Ego and Exo,
degenerate normalization denominators are marked invalid.

\begin{table*}[t]
\centering

\begin{minipage}[t]{0.68\textwidth}
\vspace{0pt}
\caption{\textbf{Dynamic disentanglement across three environments.}
Lower NOR and higher Group-IRS indicate better disentanglement.
Shading varies continuously with normalized means within
each environment and metric, darker indicates better.
Bold denotes the better result within each backbone and environment.}
\label{tab:dynamic_disentanglement}
\footnotesize
\setlength{\tabcolsep}{1pt}
\renewcommand{\arraystretch}{1.12}
\begin{tabularx}{\linewidth}{@{}lYYYY@{}}
\toprule
\textbf{Method}
& \shortstack{\textbf{Ego}\\\textbf{NOR} $\downarrow$}
& \shortstack{\textbf{Exo}\\\textbf{NOR} $\downarrow$}
& \shortstack{\textbf{Ego}\\\textbf{Group-IRS} $\uparrow$}
& \shortstack{\textbf{Exo}\\\textbf{Group-IRS} $\uparrow$} \\
\midrule
\multicolumn{5}{c}{\textbf{\textit{(a) MiniGrid}}} \\
\midrule
LeWM \cite{Maes2026LeWorldModel}
& \cellcolor{FWBlue!44.43} 0.5096 $\pm$ 0.0038
& \cellcolor{FWBlue!35.43} 1.2117 $\pm$ 0.0115
& \cellcolor{FWBlue!15.06} 0.5277 $\pm$ 0.0020
& \cellcolor{FWBlue!22.87} 0.2711 $\pm$ 0.0061 \\
\quad + CCRF (Ours)
& \cellcolor{FWBlue!100} \textbf{0.1289} $\pm$ \textbf{0.0077}
& \cellcolor{FWBlue!100} \textbf{0.1485} $\pm$ \textbf{0.0043}
& \cellcolor{FWBlue!100} \textbf{0.8671} $\pm$ \textbf{0.0076}
& \cellcolor{FWBlue!100} \textbf{0.8447} $\pm$ \textbf{0.0047} \\
DINO-WM \cite{zhou2025dinowm}
& \cellcolor{FWBlue!10} 0.7455 $\pm$ 0.0119
& \cellcolor{FWBlue!10} 1.6305 $\pm$ 0.0294
& \cellcolor{FWBlue!10} 0.5075 $\pm$ 0.0044
& \cellcolor{FWBlue!10} 0.1754 $\pm$ 0.0026 \\
\quad + CCRF (Ours)
& \cellcolor{FWBlue!49.13} \textbf{0.4774} $\pm$ \textbf{0.0118}
& \cellcolor{FWBlue!45.46} \textbf{1.0466} $\pm$ \textbf{0.0329}
& \cellcolor{FWBlue!34.35} \textbf{0.6048} $\pm$ \textbf{0.0103}
& \cellcolor{FWBlue!45.89} \textbf{0.4423} $\pm$ \textbf{0.0093} \\
\midrule
\multicolumn{5}{c}{\textbf{\textit{(b) MetaDrive}}} \\
\midrule
LeWM \cite{Maes2026LeWorldModel}
& \cellcolor{FWBlue!94.86} 0.0275 $\pm$ 0.0002
& \cellcolor{FWBlue!10} 30.5829 $\pm$ 0.7584
& \cellcolor{FWBlue!97.16} 0.9702 $\pm$ 0.0010
& \cellcolor{FWBlue!94.08} 0.1852 $\pm$ 0.0008 \\
\quad + CCRF (Ours)
& \cellcolor{FWBlue!100} \textbf{0.0165} $\pm$ \textbf{0.0012}
& \cellcolor{FWBlue!73.38} \textbf{10.7205} $\pm$ \textbf{0.8001}
& \cellcolor{FWBlue!100} \textbf{0.9756} $\pm$ \textbf{0.0004}
& \cellcolor{FWBlue!100} \textbf{0.1960} $\pm$ \textbf{0.0028} \\
DINO-WM \cite{zhou2025dinowm}
& \cellcolor{FWBlue!10} 0.2091 $\pm$ 0.0024
& \cellcolor{FWBlue!92.33} 4.7833 $\pm$ 0.0128
& \cellcolor{FWBlue!10} 0.8044 $\pm$ 0.0025
& \cellcolor{FWBlue!10} 0.0317 $\pm$ 0.0018 \\
\quad + CCRF (Ours)
& \cellcolor{FWBlue!79.81} \textbf{0.0597} $\pm$ \textbf{0.0009}
& \cellcolor{FWBlue!100} \textbf{2.3798} $\pm$ \textbf{0.0571}
& \cellcolor{FWBlue!83.81} \textbf{0.9448} $\pm$ \textbf{0.0015}
& \cellcolor{FWBlue!64.01} \textbf{0.1303} $\pm$ \textbf{0.0070} \\
\midrule
\multicolumn{5}{c}{\textbf{\textit{(c) CARLA}}} \\
\midrule
LeWM \cite{Maes2026LeWorldModel}
& \cellcolor{FWBlue!81.89} 0.1356 $\pm$ 0.0033
& \cellcolor{FWBlue!10} 7.3753 $\pm$ 0.1995
& \cellcolor{FWBlue!82.59} 0.8635 $\pm$ 0.0047
& \cellcolor{FWBlue!13.90} 0.0668 $\pm$ 0.0051 \\
\quad + CCRF (Ours)
& \cellcolor{FWBlue!96.39} \textbf{0.0890} $\pm$ \textbf{0.0055}
& \cellcolor{FWBlue!22.62} \textbf{6.5148} $\pm$ \textbf{0.2542}
& \cellcolor{FWBlue!91.36} \textbf{0.8921} $\pm$ \textbf{0.0039}
& \cellcolor{FWBlue!32.42} \textbf{0.1286} $\pm$ \textbf{0.0031} \\
DINO-WM \cite{zhou2025dinowm}
& \cellcolor{FWBlue!10} 0.3667 $\pm$ 0.0105
& \cellcolor{FWBlue!78.18} 2.7267 $\pm$ 0.0270
& \cellcolor{FWBlue!10} 0.6266 $\pm$ 0.0144
& \cellcolor{FWBlue!10} 0.0538 $\pm$ 0.0035 \\
\quad + CCRF (Ours)
& \cellcolor{FWBlue!100} \textbf{0.0774} $\pm$ \textbf{0.0019}
& \cellcolor{FWBlue!100} \textbf{1.2390} $\pm$ \textbf{0.0419}
& \cellcolor{FWBlue!100} \textbf{0.9203} $\pm$ \textbf{0.0013}
& \cellcolor{FWBlue!100} \textbf{0.3541} $\pm$ \textbf{0.0221} \\
\bottomrule
\end{tabularx}
\end{minipage}
\hfill
\begin{minipage}[t]{0.30\textwidth}
\vspace{0pt}
\centering
\includegraphics[width=0.99\linewidth]{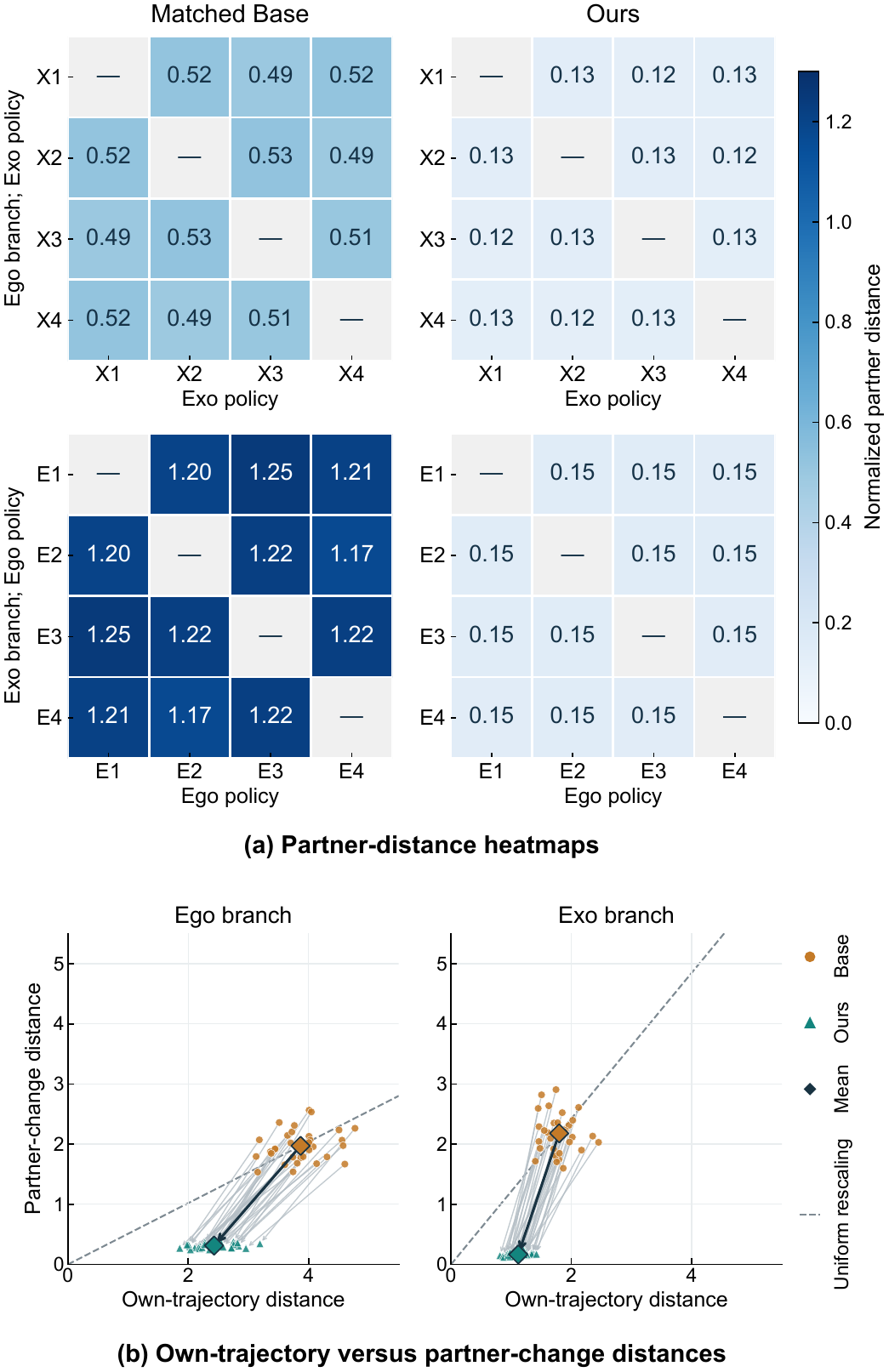}
\captionof{figure}{\textbf{Factor selectivity on MiniGrid.}
Dashed lines denote uniform rescaling, diamonds mark means.}
\label{fig:minigrid_selectivity}
\end{minipage}
\vspace{-1em}
\end{table*}

\subsection{Quantifying Dynamic Entanglement}

We quantify exogenous sensitivity using the main-effect ratio
$E_B/E_A$. As shown in Table~\ref{tab:entanglement_diagnosis},
LeWM yields ratios of 0.3773, 0.4925, and 0.0922 on MiniGrid,
MetaDrive, and CARLA before projection, while JEPA\cite{assran2023ijepa} yields
0.4217, 0.4925, and 0.0622.
These measurements show that exogenous changes contribute to
latent variation alongside agent motion, motivating representations
that selectively preserve agent-relevant changes.

Ego projection reduces relative exogenous sensitivity across
all configurations reported in the table.
For LeWM, the reductions are 94.06\%, 99.31\%, and 87.12\%
across the three environments.
For JEPA, the corresponding reductions are 8.83\%, 98.57\%,
and 86.97\%, with a smaller improvement on MiniGrid.
Each reduction is measured against the corresponding baseline.
The updated MiniGrid JEPA results use a separate 100-group
test split and an optimized projection-training protocol.
On MetaDrive, the LeWM- and JEPA-labelled rows share the same
underlying JEPA checkpoint and differ in projection training.

Fig.~\ref{fig:latent_drift} complements these aggregate
measurements by visualizing mean squared representation drift
under exogenous changes while the agent state is held fixed.
In the comparison, the original representation
exhibits substantially larger drift than either projected
representation across the displayed time steps and drift ranks.
Both projected surfaces remain within a much lower numerical range.

State supervision further lowers the ratio for LeWM on MiniGrid
and JEPA on MiniGrid and CARLA, but increases it in the
remaining configurations.
For MiniGrid JEPA, the additional improvement is modest:
the ratio decreases from 0.3844 to 0.3820.
Both main-effect energies increase, with the relative increase
in $E_A$ exceeding that in $E_B$.
Thus, a lower interference ratio need not correspond to lower
absolute exogenous drift, and state supervision does not
consistently improve factor selectivity across configurations.
These results motivate combining explicit invariance constraints
with objectives that preserve agent-relevant information.
CCRF follows this approach by constraining average and
worst-partner deviations while supervising state and motion
readouts.

\subsection{Evaluating Dynamic Disentanglement}

We evaluate bidirectional factor stability using NOR and
Group-IRS, measuring the Ego branch's response to exogenous
changes and the Exo branch's response to agent changes.
Table~\ref{tab:dynamic_disentanglement} compares LeWM and
DINO-WM with and without CCRF across three environments.
%
CCRF improves all four metrics in every backbone--environment
combination, indicating consistent gains in average relative
interference and worst-partner stability.
On MiniGrid, LeWM's Ego and Exo NOR decrease from 0.5096 and
1.2117 to 0.1289 and 0.1485, respectively, while Group-IRS
increases to 0.8671 and 0.8447.
DINO-WM also improves in both branches, with Exo NOR decreasing
from 1.6305 to 1.0466 and Exo Group-IRS increasing from 0.1754
to 0.4423. These joint improvements support the effectiveness of
counterfactual constraints beyond a single backbone and show
that reduced average interference is accompanied by greater
stability under the most disruptive partners.

Fig.~\ref{fig:minigrid_selectivity} examines the structure of these improvements. The heatmaps in panel (a) show reduced normalized partner
distances across all displayed policy pairs.
Ego distances decrease from approximately 0.49--0.53 to
0.12--0.13, while Exo distances decrease from 1.17--1.25
to approximately 0.15. The reduction therefore extends across partner combinations
rather than being concentrated in a few favorable pairs.
Panel (b) distinguishes this effect from uniform rescaling.
Although own-trajectory distances also decrease, partner-change
distances contract more strongly, placing the learned
representations below the uniform-rescaling reference lines.
This supports selective suppression of interference rather
than a uniform reduction in representation magnitude.

The driving environments reveal stronger asymmetry between
the factors. On MetaDrive, LeWM's Exo NOR decreases from 30.5829 to
10.7205, whereas Exo Group-IRS increases only from 0.1852
to 0.1960. A substantial reduction in average relative interference
thus coexists with persistent worst-partner sensitivity.
DINO-WM improves in both criteria, but its Exo branch remains
less stable than its Ego branch. On CARLA, DINO-WM's Ego and Exo NOR decrease from
0.3667 and 2.7267 to 0.0774 and 1.2390, respectively.
Exo NOR remains above one for both backbones in both driving
environments, indicating that partner variation still exceeds
own-factor variation on average.

These results support improved factor selectivity while
revealing dependence on the backbone, environment, and factor.
We next examine whether these representations support
effective decisions through closed-loop planning.

\subsection{Closed-Loop Visual Planning}

\textbf{Setup.}
We compare six representation configurations with CEM,
Beam Search, and our planner across three environments.
Each run evaluates the same 150 episodes per environment
with a 50-step budget.
Success requires reaching the target in MiniGrid or entering
a 5\,m goal region in MetaDrive and CARLA, where goal images are taken 25 recorded steps after the start.
We report mean SR $\pm$ sample SD. Our planner uses five readout seeds in MiniGrid and MetaDrive and three in CARLA. Search baselines use three configurations.
The MetaDrive variant uses a supervised action prior and
semantic costs with CEM optimization. Methods share environment evaluation criteria but differ in supervision, control, and compute, so comparisons assess
complete planning systems.

\textbf{Results Analysis.}
Our planner achieves higher mean SR than both search baselines
for every representation in all three environments as in Table~\ref{tab:planning_sr}.
The largest differences occur in MiniGrid, where Full latent
achieves 78.80\% with our planner, compared with 13.56\% for CEM
and 14.15\% for Beam Search.
With Full + SEX routing, these values become 82.40\%, 19.20\%,
and 18.27\%, respectively.
The strong Full baseline indicates that the task prediction
and planning interface accounts for a substantial part of
the observed performance difference.
Factor routing provides a further 3.60 percentage point
improvement in our planner.

The contribution of factor information depends on how it is
used. In MiniGrid, directly using SEX yields 78.93\% with our
planner, close to Full latent and below CCRF SEX at
81.20\%. Thus, the stronger disentanglement measured in the preceding
section does not translate monotonically into higher SR.
Full + SEX achieves the highest mean, supporting the use of
factors as semantic corrections to the Full prediction pathway.
Using SE or Ego alone yields lower means of 76.80\% and 75.33\%,
although these comparisons do not by themselves isolate the
contribution of individual factors.

The additional benefit of routing is smaller in the driving
environments. On MetaDrive, Full latent, Full + SEX, and
CCRF SEX all achieve 99.33\% with our planner.
On CARLA, Full + SEX reaches 98.00\%, compared with 97.56\%
for Full latent, a difference of 0.44 percentage points.
Routing also fails to improve CEM or Beam Search consistently
across these environments. Together, these results indicate that factor utility depends
on the planning interface and task, rather than following
directly from reduced representational interference.

\begin{table}[t]
\centering
\caption{\textbf{Closed-loop planning success rates.}
Values are mean SR (\%) $\pm$ sample SD over runs of
150 episodes each.
SEX combines Scene, Ego, and Exo. SE uses Scene and Ego.
Shading is normalized within each planner and environment.
Darker indicates higher SR; bold marks the best mean.}
\label{tab:planning_sr}
\small
\setlength{\tabcolsep}{3pt}
\renewcommand{\arraystretch}{1.12}
\begin{tabularx}{\linewidth}{@{}lYYY@{}}
\toprule
\textbf{Representation}
& \textbf{CEM}
& \textbf{Beam Search}
& \textbf{Ours} \\
\midrule
\multicolumn{4}{c}{\textbf{\textit{(a) MiniGrid}}} \\
\midrule
LeWM Full latent
& \cellcolor{FWSRDark!33.56!FWSRLight} 13.56 $\pm$ 1.02
& \cellcolor{FWSRDark!57.43!FWSRLight} 14.15 $\pm$ 1.32
& \cellcolor{FWSRDark!54.17!FWSRLight} 78.80 $\pm$ 3.18 \\
Full + SEX
& \cellcolor{FWSRDark!100!FWSRLight} \textbf{19.20} $\pm$ 3.31
& \cellcolor{FWSRDark!100!FWSRLight} \textbf{18.27} $\pm$ 1.46
& \cellcolor{FWSRDark!100!FWSRLight} \textbf{82.40} $\pm$ 1.38 \\
CCRF SEX
& \cellcolor{FWSRDark!46.64!FWSRLight} 14.67 $\pm$ 2.00
& \cellcolor{FWSRDark!55.88!FWSRLight} 14.00 $\pm$ 0.67
& \cellcolor{FWSRDark!84.72!FWSRLight} 81.20 $\pm$ 3.18 \\
SEX
& \cellcolor{FWSRDark!10!FWSRLight} 11.56 $\pm$ 1.02
& \cellcolor{FWSRDark!14.55!FWSRLight} 10.00 $\pm$ 0.67
& \cellcolor{FWSRDark!55.83!FWSRLight} 78.93 $\pm$ 4.15 \\
SE
& \cellcolor{FWSRDark!12.59!FWSRLight} 11.78 $\pm$ 2.14
& \cellcolor{FWSRDark!28.29!FWSRLight} 11.33 $\pm$ 0.00
& \cellcolor{FWSRDark!28.71!FWSRLight} 76.80 $\pm$ 5.11 \\
Ego
& \cellcolor{FWSRDark!23.08!FWSRLight} 12.67 $\pm$ 1.76
& \cellcolor{FWSRDark!10!FWSRLight} 9.56 $\pm$ 1.02
& \cellcolor{FWSRDark!10!FWSRLight} 75.33 $\pm$ 3.86 \\
\midrule
\multicolumn{4}{c}{\textbf{\textit{(b) MetaDrive}}} \\
\midrule
LeWM Full latent
& \cellcolor{FWSRDark!96.70!FWSRLight} 89.56 $\pm$ 1.39
& \cellcolor{FWSRDark!100!FWSRLight} \textbf{78.67} $\pm$ 1.26
& \cellcolor{FWSRDark!100!FWSRLight} \textbf{99.33} $\pm$ 0.00 \\
Full + SEX
& \cellcolor{FWSRDark!10!FWSRLight} 83.78 $\pm$ 1.02
& \cellcolor{FWSRDark!10!FWSRLight} 47.11 $\pm$ 1.54
& \cellcolor{FWSRDark!100!FWSRLight} \textbf{99.33} $\pm$ 0.00 \\
CCRF SEX
& \cellcolor{FWSRDark!100!FWSRLight} \textbf{89.78} $\pm$ 1.02
& \cellcolor{FWSRDark!43.59!FWSRLight} 58.89 $\pm$ 2.52
& \cellcolor{FWSRDark!100!FWSRLight} \textbf{99.33} $\pm$ 0.00 \\
SEX
& \cellcolor{FWSRDark!86.65!FWSRLight} 88.89 $\pm$ 2.34
& \cellcolor{FWSRDark!32.81!FWSRLight} 55.11 $\pm$ 1.39
& \cellcolor{FWSRDark!61.29!FWSRLight} 98.93 $\pm$ 0.37 \\
SE
& \cellcolor{FWSRDark!76.60!FWSRLight} 88.22 $\pm$ 0.77
& \cellcolor{FWSRDark!32.19!FWSRLight} 54.89 $\pm$ 1.39
& \cellcolor{FWSRDark!48.71!FWSRLight} 98.80 $\pm$ 0.56 \\
Ego
& \cellcolor{FWSRDark!86.65!FWSRLight} 88.89 $\pm$ 1.02
& \cellcolor{FWSRDark!25.86!FWSRLight} 52.67 $\pm$ 1.15
& \cellcolor{FWSRDark!10!FWSRLight} 98.40 $\pm$ 0.76 \\
\midrule
\multicolumn{4}{c}{\textbf{\textit{(c) CARLA}}} \\
\midrule
LeWM Full latent
& \cellcolor{FWSRDark!48.65!FWSRLight} 91.78 $\pm$ 1.92
& \cellcolor{FWSRDark!100!FWSRLight} \textbf{92.22} $\pm$ 1.39
& \cellcolor{FWSRDark!74.62!FWSRLight} 97.56 $\pm$ 1.39 \\
Full + SEX
& \cellcolor{FWSRDark!22.69!FWSRLight} 91.33 $\pm$ 0.67
& \cellcolor{FWSRDark!10!FWSRLight} 89.78 $\pm$ 0.38
& \cellcolor{FWSRDark!100!FWSRLight} \textbf{98.00} $\pm$ 0.67 \\
CCRF SEX
& \cellcolor{FWSRDark!10!FWSRLight} 91.11 $\pm$ 1.02
& \cellcolor{FWSRDark!67.17!FWSRLight} 91.33 $\pm$ 0.67
& \cellcolor{FWSRDark!74.62!FWSRLight} 97.56 $\pm$ 0.38 \\
SEX
& \cellcolor{FWSRDark!100!FWSRLight} \textbf{92.67} $\pm$ 2.31
& \cellcolor{FWSRDark!42.83!FWSRLight} 90.67 $\pm$ 0.00
& \cellcolor{FWSRDark!87.31!FWSRLight} 97.78 $\pm$ 0.38 \\
SE
& \cellcolor{FWSRDark!74.04!FWSRLight} 92.22 $\pm$ 1.68
& \cellcolor{FWSRDark!50.94!FWSRLight} 90.89 $\pm$ 0.38
& \cellcolor{FWSRDark!61.35!FWSRLight} 97.33 $\pm$ 0.00 \\
Ego
& \cellcolor{FWSRDark!48.65!FWSRLight} 91.78 $\pm$ 1.02
& \cellcolor{FWSRDark!10!FWSRLight} 89.78 $\pm$ 0.38
& \cellcolor{FWSRDark!10!FWSRLight} 96.44 $\pm$ 0.38 \\
\bottomrule
\end{tabularx}
\vspace{-1em}
\end{table}


\subsection{Task Adaptation for Continuous Control}
\label{sec:continuous_control}

\textbf{Setup.}
We evaluate task-adapted FactorWorld systems on Two-Room,
Reacher, Push-T, and Cube, covering navigation and manipulation.
Starts and goal images are drawn from offline trajectories,
with goal offsets of 100 steps for Two-Room and 25 otherwise,
and interaction budgets of 150 and 50 steps, respectively.
Success follows the environment criteria. The systems combine frozen visual representations and semantic readouts with task-specific controllers trained using additional
state or action supervision, while execution uses images and
action history without ground-truth state access.
After validation-based selection, all four tasks use independent
confirmation sets, with 50 episodes for Two-Room and Reacher
and 100 for Push-T and Cube. We report mean SR and sample SD across three training repeats
sharing the same confirmation episodes within each task. Controller details and full protocols are provided in the Appendix.

\begin{figure}[htbp]
\centering
\begin{minipage}[t]{0.51\linewidth}
  \vspace{0pt}
  \centering
  \includegraphics[width=\linewidth]
    {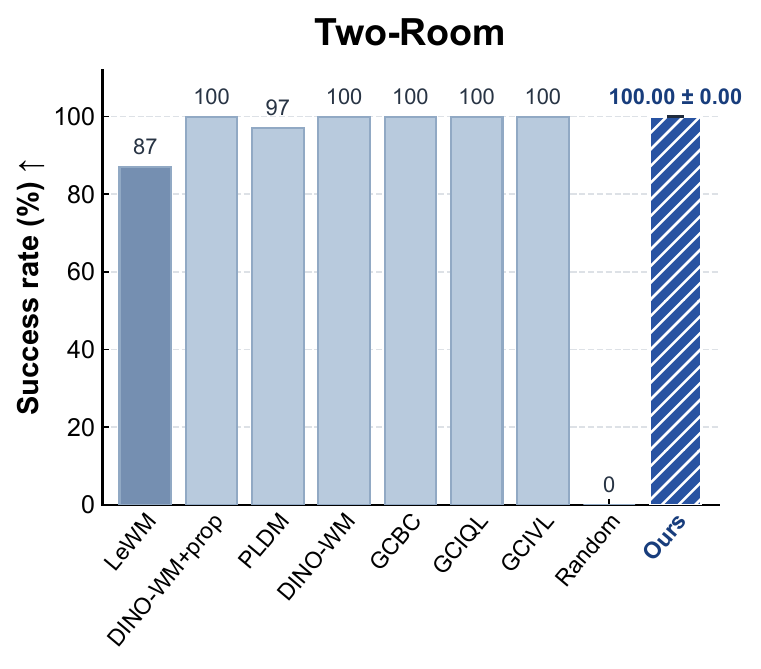}
\end{minipage}\hfill
\begin{minipage}[t]{0.48\linewidth}
  \vspace{0pt}
  \centering
  \includegraphics[width=\linewidth]
    {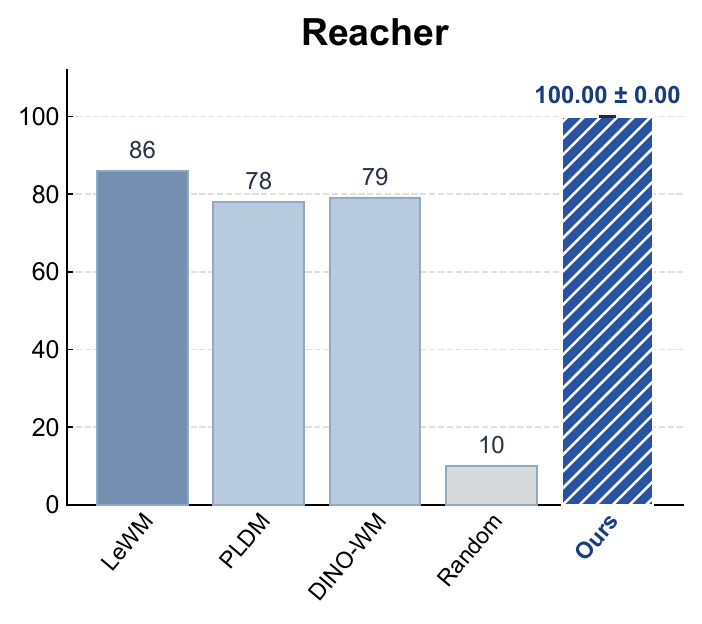}
\end{minipage}
\par\vspace{0pt}
\begin{minipage}[t]{0.50\linewidth}
  \vspace{0pt}
  \centering
  \includegraphics[width=\linewidth]
    {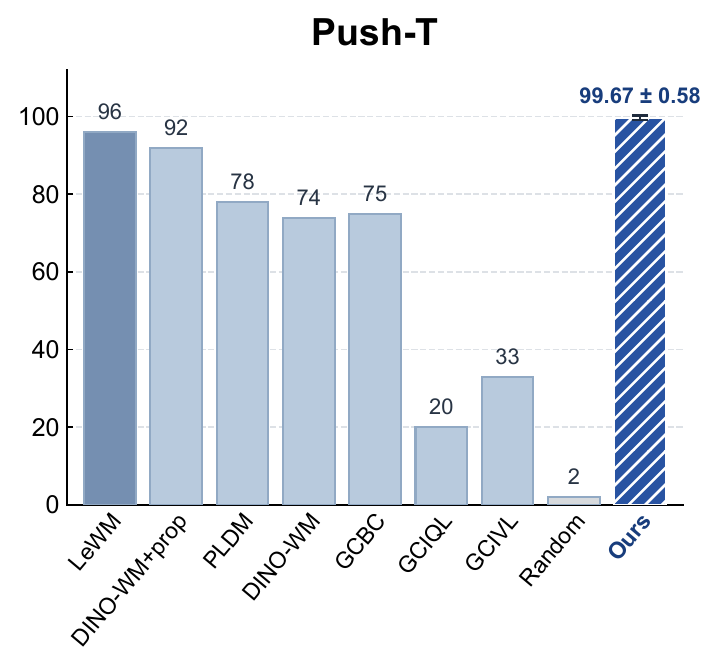}
\end{minipage}\hfill
\begin{minipage}[t]{0.49\linewidth}
  \vspace{0pt}
  \centering
  \includegraphics[width=\linewidth]
    {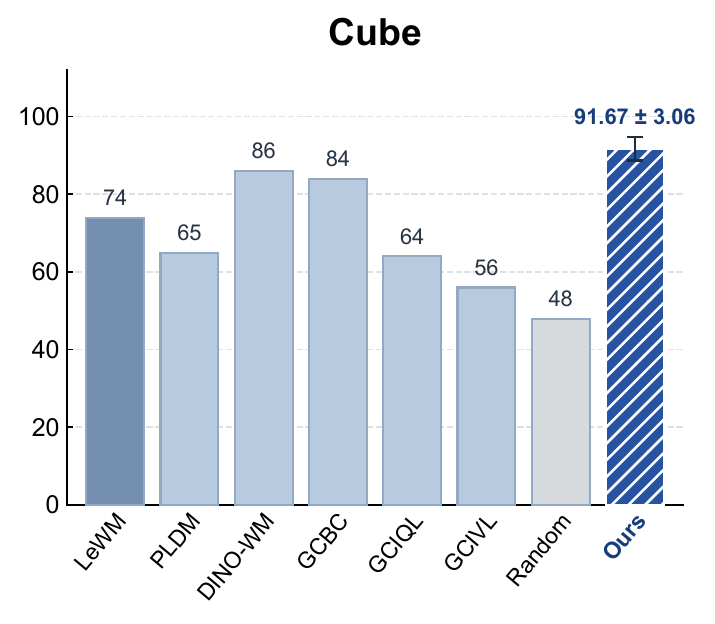}
\end{minipage}
\caption{\textbf{Success rates on four continuous control tasks.}
Hatched bars denote task-adapted FactorWorld systems.
Baseline scores are reported by
LeWM~\cite{Maes2026LeWorldModel}.}
\label{fig:p4}
\vspace{-0.5em}
\end{figure}


\textbf{Results Analysis.}
Fig.~\ref{fig:p4} summarizes the results.
Two-Room and Reacher achieve 100.00\% $\pm$ 0.00 across
three repeats, while Push-T reaches 99.67\% $\pm$ 0.58,
exceeding the reported LeWM score by 3.67 percentage points.
These results are consistent with the benefit of connecting
visual representations to explicit control quantities.
In Two-Room, estimated poses and image-derived obstacles
support search over feasible paths.
In Reacher, spatially corrected joint estimates provide
feedback in the coordinates used by the controller.
Both interfaces connect visual observations to actionable
state estimates, reducing reliance on latent distance as
a proxy for task progress.
Push-T further illustrates the importance of aligning the
planning objective with task completion.
Calibrated semantic readouts expose object position and
orientation to learned contact dynamics, while the search
cost incorporates success tolerances and rewards reaching
the target region anywhere along the predicted trajectory.
This allows the planner to recognize successful intermediate
states rather than requiring the final predicted state to
match the goal exactly.
The high SR across training repeats supports the effectiveness
of this combined perception and contact-planning interface.
On Cube, our task-adapted semantic routing planner achieves
91.67\% $\pm$ 3.06, exceeding the published LeWM and DINO-WM
scores by 17.67 and 5.67 percentage points, respectively.
The planner uses visual semantic predictions both to define
an object-centered goal cost and to generate action proposals.
This connects task state estimation with search initialization
and goal evaluation. On the same validation set with a fixed readout seed, adding proposal initialization increases SR from 74\% to 91\%,
supporting the contribution of guided search within our planner.

Overall, these results highlight the value of using semantic
information throughout decision-making, from state estimation
and feedback to trajectory scoring and search initialization.
The shared design principle is to connect visual factors to
the quantities and operations required by each task.

%% file: sec/5_conclusion.tex
\section{Conclusion}

Mixed agent and exogenous dynamics can make latent variation
an unreliable measure of task progress in visual world models.
We presented FactorWorld to address this challenge by connecting
counterfactual dynamic disentanglement with semantic routing
for planning.
Bidirectional counterfactual pairs quantify cross-factor
interference, while CCRF combines stability constraints with
state and motion supervision to learn selective representations.
Semantic routing uses these factors to refine task predictions
through explicit decision interfaces.
Experiments show improved factor stability across two backbones
and three environments, a MiniGrid success increase from
78.80\% to 82.40\%, and mean success rates of 91.67--100\%
across four continuous control adaptations. These results support connecting dynamic disentanglement with explicit planning interfaces.

Future work will explore how task objectives and active
interventions can jointly guide world model revision,
distinguishing representation errors from changes in
environment dynamics and determining which changes to
ignore, model, or anticipate for planning.

%% file: sec/X_suppl.tex
\clearpage
\setcounter{page}{1}
\maketitlesupplementary

\section{Experimental Details}
\label{app:protocols}

\subsection{Data Construction and Splits}
\label{app:data}

For each scene, we cross four Ego trajectories with four Exo
trajectories to obtain 16 branches. We align observations by
time so that we can vary either trajectory while holding the
other fixed. All branches from a scene belong to the same split.
Table~\ref{tab:app_data_splits} summarizes the observation
settings, motion controls, and splits.

In MiniGrid, we sample six aligned time points from each
120-step trajectory. Factor training uses 12 of the 16
combinations per training map. We evaluate factor stability
on all 16 combinations per validation map and reserve the
test maps for separate generalization experiments.
We train a separate factor model for each environment.
For continuous control, we sample starts and goal images from
offline trajectories. Goal images are taken 100 recorded steps
after the start in Two-Room and 25 steps after the start in
Reacher, Push-T, and Cube. Confirmation episodes are excluded
from adaptation training and validation. 

\noindent
\begin{minipage}{\columnwidth}
\centering
\captionof{table}{\textbf{Counterfactual data construction and splits.}
Splits count maps in MiniGrid and scene groups in MetaDrive and CARLA.
Each map or scene group contains 16 trajectory combinations.}
\label{tab:app_data_splits}
\footnotesize
\setlength{\tabcolsep}{2pt}
\renewcommand{\arraystretch}{1.03}
\begin{tabularx}{\linewidth}
  {@{}l *{3}{>{\raggedright\arraybackslash}X}@{}}
\toprule
\textbf{Setting} & \textbf{MiniGrid} & \textbf{MetaDrive} & \textbf{CARLA} \\
\midrule
Scene
& $11\times11$ grid
& Road scenes
& Town04 \\
RGB view
& Full map, top-down
& Ego-centered, top-down
& Scene, top-down \\
Ego actions
& Turn left/right, move forward, stay still
& Steering, throttle, braking
& Steering, throttle \\
Exo entities
& Independently moving blue ball
& Traffic vehicles
& Traffic vehicles \\
Exo control
& Open-loop random walk
& Seeded open-loop traffic control
& Deterministic open-loop route following \\
\midrule
Total      & 320 & 64 & 64 \\
Training   & 224 & 48 & 48 \\
Validation &  32 & 16 &  8 \\
Test       &  64 & -- &  8 \\
\bottomrule
\end{tabularx}
\end{minipage}
\par

\begin{table}[htbp]
\centering
\caption{\textbf{MiniGrid architecture and training settings.}
The visual encoder remains frozen throughout.
During readout training, the factorizer is also frozen.}
\label{tab:app_training}
\footnotesize
\setlength{\tabcolsep}{4pt}
\renewcommand{\arraystretch}{1.08}
\begin{tabularx}{\linewidth}
{@{}l>{\raggedright\arraybackslash}X@{}}
\toprule
\textbf{Setting} & \textbf{Value} \\
\midrule
\multicolumn{2}{c}{\textit{Factorization}} \\
\midrule
Full latent dimension & 192 \\
Scene / Ego / Exo dimensions & 64 / 64 / 64 \\
MLP hidden dimension & 256 \\
Normalization / activation & LayerNorm / GELU \\
Training epochs & 30 \\
Batch size & 16 scene groups \\
Partner-invariance weight & 4.0 \\
Worst-partner weight & 4.5 \\
Training seeds & 6101, 6102, 6103 \\
\midrule
\multicolumn{2}{c}{\textit{Task Readouts}} \\
\midrule
Training data & 98,000 images with spatial labels \\
Learning rate & $5\times10^{-4}$ \\
Readout seeds & 3301--3305 \\
Residual regularization & $10^{-3}$ for residual adapters;
none for factor-only heads \\
Adapter output initialization & Zero \\
Residual checkpoint epochs & $\{1,3,10\}$ \\
Factor-only checkpoint epochs & $\{10,30,60\}$ \\
Selection criterion & SR on 50 validation episodes;
earlier epoch breaks ties \\
\bottomrule
\end{tabularx}
\vspace{-1em}
\end{table}

\begin{table*}[htbp]
\centering
\caption{\textbf{Evaluation protocols.}
Episode counts and step budgets are given per run.
The repeat counts are for our planner.}
\label{tab:app_evaluation}
\small
\begin{tabularx}{\textwidth}{
  @{}l l c c c >{\raggedright\arraybackslash}X@{}
}
\toprule
\textbf{Setting} & \textbf{Task} & \textbf{Episodes}
& \textbf{Step budget} & \textbf{Repeats}
& \textbf{Success criterion} \\
\midrule
Navigation & MiniGrid  & 150 & 50  & 5 & Reach the target position \\
           & MetaDrive & 150 & 50  & 5 & Enter the 5\,m goal region \\
           & CARLA     & 150 & 50  & 3 & Enter the 5\,m goal region \\
\midrule
Continuous control & Two-Room & 50  & 150 & 3 & Environment-defined success \\
                   & Reacher  & 50  & 50  & 3 & Environment-defined success \\
                   & Push-T   & 100 & 50  & 3 & Environment-defined success \\
                   & Cube     & 100 & 50  & 3 & Environment-defined success \\
\bottomrule
\end{tabularx}
\vspace{-0.5em}
\end{table*}

\begin{table*}[htbp]
\centering
\caption{\textbf{Baseline and input configurations.}
The planning rows specify the input to each configuration.
For MiniGrid semantic routing, Full + SEX retains the frozen
Full readout and adds factor residuals to its task predictions.}
\label{tab:app_baseline_configs}
\small
\begin{tabularx}{\textwidth}{
  @{}l l >{\raggedright\arraybackslash}X@{}
}
\toprule
\textbf{Evaluation} & \textbf{Configuration} & \textbf{Definition} \\
\midrule
Interference diagnosis
& Full latent
& Original representation from a frozen visual encoder. \\
& Ego projection
& Learned Ego projection of the frozen representation. \\
& Ego projection + state supervision
& Ego projection trained with additional state supervision. \\
\midrule
Factor stability
& Matched direct projection
& MiniGrid LeWM control that maps the Full latent directly to
  three factor codes, with a parameter count matched to CCRF. \\
& CCRF
& Extracts Ego and Exo codes from the residual after
  subtracting the decoded Scene code from the Full latent. \\
\midrule
Closed-loop planning
& Full & Original visual representation. \\
& Full + SEX & Full representation and all three factors. \\
& CCRF SEX & Scene, Ego, and Exo from the basic CCRF model. \\
& SEX & Scene, Ego, and Exo from the final factor model. \\
& SE & Scene and Ego from the final factor model. \\
& Ego & Ego from the final factor model. \\
\bottomrule
\end{tabularx}
\vspace{-1em}
\end{table*}

\subsection{Architecture and Training}
\label{app:training}

Training proceeds in two stages.
We first train CCRF on frozen visual features, then freeze
both the encoder and factorizer to train task readouts.
Table~\ref{tab:app_training} summarizes the MiniGrid
architecture and training settings.

CCRF extracts the dynamic factors after subtracting the
decoded scene code from the full representation.
For semantic routing, Ego corrects pose predictions,
while Scene, Ego, and Exo jointly correct goal and
occupancy predictions.
Zero-initialized adapter output layers preserve the
base predictions at the start of training.
Factor-only readouts instead predict task quantities
without a Full pathway.
The base readout also remains frozen during adapter training.

\subsection{Evaluation}
\label{app:evaluation}

Table~\ref{tab:app_evaluation} summarizes the evaluation
protocols. All methods evaluated on the same task use the
same start--goal pairs across runs. For MetaDrive and CARLA,
the goal image is taken 25 recorded steps after the start.
Driving success measures goal reaching regardless of
whether the agent stops at the goal or avoids collisions.

For navigation, our planner varies the readout training
seed. Search baselines generally use three configurations,
varying the search seed, factor model, or readout seed.
The MiniGrid Full + SEX search evaluation uses five
readout seeds. For continuous control, repeats vary
readout training in Two-Room and Cube, and also vary
task adaptation training in Reacher and Push-T.
The Cube repeats share a frozen world model and action
proposal network.




\subsection{Baselines and Supervision}
\label{app:baselines}

Table~\ref{tab:app_baseline_configs} defines the configurations
used to diagnose interference, assess factor stability, and
evaluate closed-loop planning.
In the MetaDrive interference diagnosis, the groups labeled
LeWM and JEPA share a single JEPA checkpoint and use different
projection training configurations. Factor stability is
measured on explicit Ego and Exo codes. Applying the same
directional comparisons to the Full latent measures sensitivity
of the shared representation.

Training uses state, displacement, action, or occupancy labels
according to the objective. At deployment, the policy uses
observations, a goal image, and action history where applicable.
The evaluator uses ground-truth environment state for episode
initialization and success checks.

CEM, Beam Search, and the task-adapted planners differ in
control candidates, supervision, feedback frequency, and
compute budget. Their results compare complete planning systems.
Continuous control baselines are quoted from published results
under their original protocols. Adaptation-specific supervision
and controller settings are detailed in
Appendix~\ref{app:adaptations}.



\section{Metrics}
\label{app:metrics}

\subsection{Main-Effect Analysis}
\label{app:main_effects}

For each scene and aligned time step, let $z_{ab}$ denote the
representation under motion settings $a,b\in\{0,1\}$, where
$a$ enables Ego motion and $b$ enables Exo motion.
Using the branch with both motions disabled as a reference,
we compute the Ego and Exo main-effect energies as
\begin{equation}
\begin{aligned}
E_A &= \mathbb{E}\!\left[\|z_{10}-z_{00}\|_2^2\right],\\
E_B &= \mathbb{E}\!\left[\|z_{01}-z_{00}\|_2^2\right].
\end{aligned}
\end{equation}
The expectations average over evaluated scenes and time steps.
The ratio $E_B/E_A$ is computed after averaging the two energies.
For a projected representation with ratio $r$ and its
corresponding baseline with ratio $r_0$, the relative
reduction is $100(1-r/r_0)\%$.
The ratio is invariant to uniform rescaling of the
representation. We report both energies alongside the ratio
because a lower ratio can reflect weaker Exo sensitivity,
stronger Ego sensitivity, or both.

\subsection{NOR and Group-IRS}
\label{app:factor_metrics}

We evaluate each dynamic branch using all 16 trajectory
combinations in each scene--time group. For the Ego branch,
Ego trajectories are the own trajectories and Exo trajectories
are the partners. The roles are reversed for the Exo branch.

Let $D_{\mathrm{partner}}$ be the mean Euclidean distance
between codes that share an own trajectory and differ in
their partner. Let $D_{\mathrm{own}}$ be the corresponding
mean distance when the partner is fixed and the own trajectory
changes. Each unordered pair is counted once.
NOR is $D_{\mathrm{partner}}/D_{\mathrm{own}}$,
with both distances pooled over the evaluation groups
before division.

Group-IRS is computed separately for each scene--time group.
For each own trajectory, we find the largest distance from
its mean code across partners. We average these maxima over
own trajectories and divide by the largest distance from
the overall group mean. The group score is one minus this
ratio, and the final score averages over groups.
Degenerate normalization denominators are flagged as invalid.
We compute the metrics for each training seed before
aggregating across seeds.

\subsection{Drift and Selectivity}
\label{app:drift_selectivity}

\paragraph{Drift surfaces.}
To visualize Exo-induced changes with Ego motion enabled,
we compare the joint-motion branch with the Ego-only branch.
For latent coordinate $k$ at time $t$, the plotted statistic
before smoothing is
\begin{equation}
d_{t,k}=\mathbb{E}_{s}\!\left[
  \bigl(z_{11}^{(s,t,k)}-z_{10}^{(s,t,k)}\bigr)^2
\right],
\end{equation}
where $s$ indexes evaluation scenes. Within each
representation, coordinates are ranked by their mean drift
over time, and the 48 highest-ranked coordinates are shown.
The projection panels use a $3\times3$ moving average for
display. Colors are normalized within each panel, and the
vertical axes show the magnitude of the drift.

\paragraph{Partner distances and uniform rescaling.}
In the MiniGrid partner-distance heatmaps, each off-diagonal
entry averages the distance for a fixed partner pair over
validation maps, time steps, and own trajectories. We divide
each entry by the mean own-trajectory distance for the same
branch and seed, then average across three seeds. Averaging
the entries over distinct partner pairs recovers mean NOR.

In the accompanying scatter plots, each point shows a
validation map's mean own and partner distances, averaged
across seeds. The dashed line passes through the origin and
the matched baseline mean, preserving its partner-to-own
distance ratio. A CCRF mean below the line indicates a
lower ratio than the matched baseline.

\section{Disentanglement Analysis}
\label{app:disentanglement}

The following diagnostics examine the early additive CCRF and its
matched direct-projection control. Their shared reconstruction
setup allows us to test which states are readable from the Scene
code, how removing a decoded branch affects state prediction, and
whether codes from observed trajectories can be recombined on
held-out maps. The selected CCRF was trained with reconstruction
and composition losses disabled.

\subsection{Matched Controls and Scene Leakage}
\label{app:scene_leakage}

We compare an early CCRF with a parameter-matched direct
projection control on MiniGrid. CCRF encodes the dynamic
factors from a scene residual and uses an additive decoder.
The control projects the full latent directly into three
64-dimensional codes and decodes their concatenation.
The models have 399,762 and 398,610 parameters, respectively,
and use the same frozen features, training objectives, and
paired seeds.

\begin{table}[htbp]
\centering
\small
\caption{\textbf{Dynamic state readout from the Scene code.}
Values are standardized probe MSE, reported as mean
$\pm$ sample SD over three model seeds on 64 held-out maps.
Lower error indicates greater state predictability.}
\label{tab:app_scene_leakage}
\begin{tabular*}{\linewidth}{@{\extracolsep{\fill}}lcc@{}}
\toprule
\textbf{Probe target} & \textbf{Matched control} & \textbf{Early CCRF} \\
\midrule
Ego state & $0.487\pm0.023$ & $0.335\pm0.024$ \\
Exo state & $0.597\pm0.008$ & $0.555\pm0.031$ \\
\bottomrule
\end{tabular*}
\end{table}

To test whether the Scene code carries dynamic information,
we fit linear state probes on the observed combinations of
the training maps and evaluate them on all combinations of
the 64 held-out maps. Lower probe error means the dynamic
state is easier to recover from the Scene code.
Table~\ref{tab:app_scene_leakage} shows that both states
are more predictable from the early CCRF Scene code than
from the matched control. Scene residualization can
therefore leave dynamic information in the Scene branch.

\begin{figure*}[!htbp]
  \centering
  \includegraphics[width=0.98\linewidth]{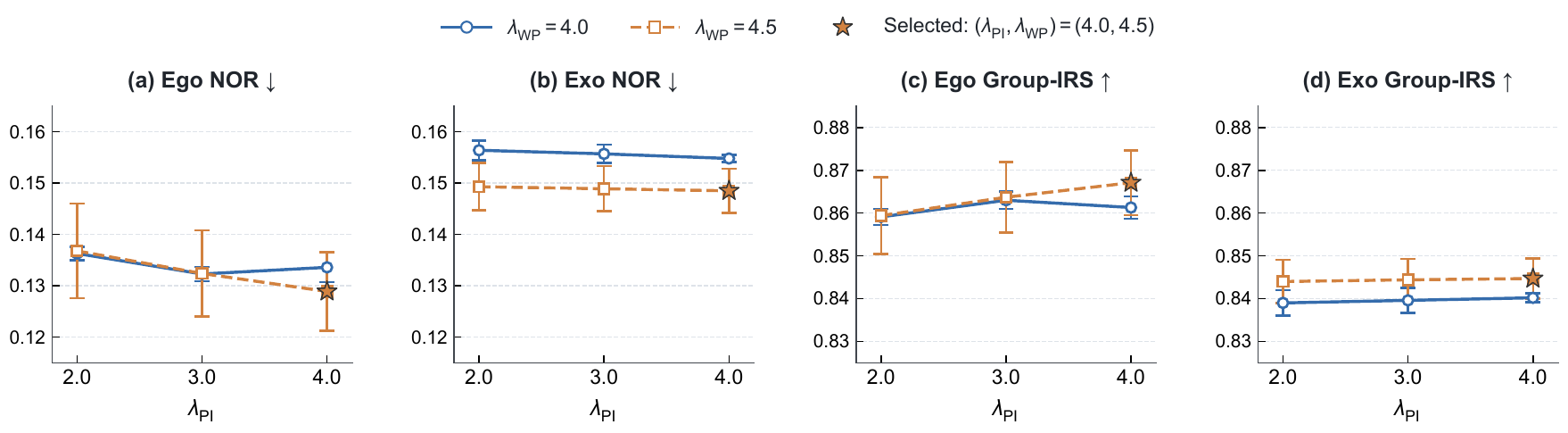}
  \caption{\textbf{Sensitivity to invariance loss weights on MiniGrid.}
  Points and error bars show the mean and sample standard deviation
  across three model seeds on the validation split.
  The selected configuration uses
  $\lambda_{\mathrm{PI}}=4.0$ and $\lambda_{\mathrm{WP}}=4.5$.}
  \label{tab:app_loss_sensitivity}
\vspace{-1em}
\end{figure*}

\subsection{Information Retention}
\label{app:information_retention}

To test what each dynamic branch contributes, we zero its
decoded contribution in an early additive CCRF variant while
keeping the model and linear probes fixed. The probes are
fitted on full latents from the training maps and applied to
the resulting composed latents on validation maps.

\begin{figure}[htbp]
  \centering
  \includegraphics[width=0.98\linewidth]{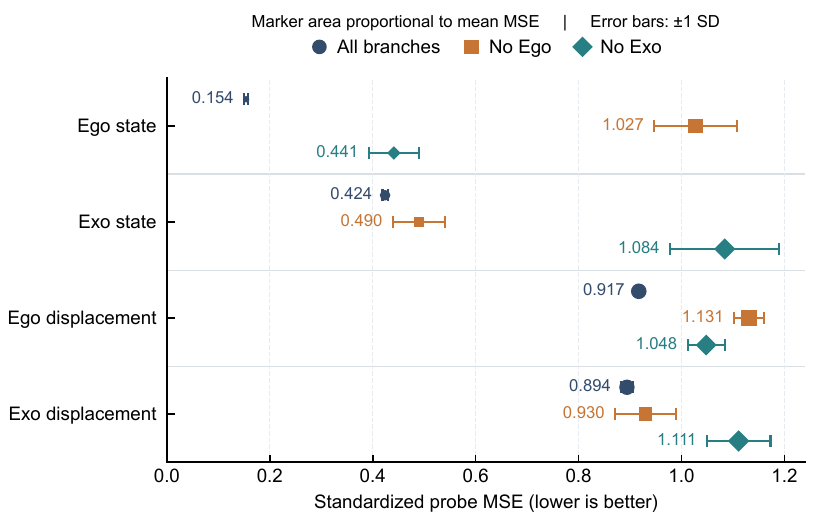}
  \caption{\textbf{Branch removal in an early additive CCRF on MiniGrid.}
Removing either branch increases probe error most for its own state
and displacement. Marker position and area encode mean standardized
MSE on held-out trajectory combinations from 32 validation maps.
Error bars show one sample standard deviation across three model seeds.}
  \label{tab:2}
  \vspace{-1em}
\end{figure}

In Fig.~\ref{tab:2}, removing either
branch raises the corresponding state and displacement
errors. The larger changes for state readouts indicate
that this intervention most clearly exposes the branches'
contribution to state prediction. State errors also rise
when the other branch is removed, showing residual overlap
in the information used by the probes.


\subsection{Compositional Generalization}
\label{app:compositional_generalization}

We use the early CCRF and matched control from the
Scene leakage comparison to test whether their codes
can reconstruct the frozen encoder latent for unseen
trajectory combinations. For each map and time step,
we estimate factor codes from the 12 observed combinations
and compose them to reconstruct the four held-out
combinations. We evaluate this procedure on 64 held-out maps,
so it tests new combinations on new maps.

Table~\ref{tab:app_composition} shows that early CCRF
reduces direct reconstruction error but increases error
on held-out combinations. The paired map bootstrap
gives a 95\% confidence interval of $[0.0156,0.0250]$
for the increase in composed MSE. This result separates
the ability to reconstruct an individual observation
from the ability to recombine factors on unseen inputs.
It characterizes the early reconstruction-trained
models. The selected MiniGrid CCRF is evaluated for
factor stability and downstream planning.

\begin{table}[htbp]
\centering
\small
\caption{\textbf{Latent reconstruction on held-out maps.}
Values are MSE against the frozen encoder latent, reported
as mean $\pm$ sample SD over three model seeds.
Lower is better.}
\label{tab:app_composition}
\begin{tabular}{@{}lcc@{}}
\toprule
\textbf{Reconstruction} & \textbf{Matched control} & \textbf{Early CCRF} \\
\midrule
Direct                & $0.14056\pm0.00233$ & $0.13213\pm0.00185$ \\
Held-out combination  & $0.17234\pm0.00225$ & $0.19256\pm0.00239$ \\
\bottomrule
\end{tabular}
\vspace{-1em}
\end{table}

\subsection{Loss Sensitivity}
\label{app:loss_sensitivity}

We vary the weights on partner invariance and worst-partner
deviation while holding the other loss weights fixed.
The own-variation weight is 1.0, and the reconstruction and
composition weights are zero. Fig.~\ref{tab:app_loss_sensitivity}
reports stability on the 32 MiniGrid validation maps.
At $\lambda_{\mathrm{WP}}=4.5$, increasing
$\lambda_{\mathrm{PI}}$ from 2.0 to 4.0 improves all four
mean scores, with a larger change for Ego than for Exo.
The selected setting, $(4.0,4.5)$, gives the best mean on all
four metrics among the configurations evaluated here.

\vspace{-1em}
\paragraph{Cross-factor leakage in the selected model.}
We probe the same three MiniGrid CCRF checkpoints used for the
factor stability evaluation. Given the Scene code and the branch's
own ground-truth state, we measure how much adding the branch code
reduces probe MSE for the partner's state on the 32 validation maps.
Adding the Ego code reduces Exo state prediction error by 13.72\%
with a ridge probe and 10.02\% with an MLP probe. In the reverse
direction, adding the Exo code reduces Ego state prediction error
by 0.88\% and 1.72\%, respectively. Thus, the selected model's
lower partner sensitivity coexists with residual cross-factor
information, most clearly in the Ego code.


\section{Planning Analysis}
\label{app:planning}

\subsection{Stability and Planning Utility}
\label{app:stability_utility}

An earlier MiniGrid experiment directly tests whether robustness
to exogenous changes predicts the usefulness of a planning cost.
For each of 50 counterfactual groups, we hold the agent state,
goal image, and 300 candidate action sequences fixed while
changing the exogenous context. Each sequence spans ten steps.
We measure agreement of the selected first action, Spearman
correlation between candidate rankings, and the fraction of
candidate pairs whose order reverses. A separate 50-episode CEM
evaluation measures closed-loop success for the same four costs.

\begin{table}[htbp]
\centering
\caption{\textbf{Counterfactual ranking stability and closed-loop
success on MiniGrid.} Agreement and flips are percentages.
State cost measures distance to the goal using a learned
position readout.}
\label{tab:app_stability_utility}
\small
\setlength{\tabcolsep}{3pt}
\begin{tabular*}{\linewidth}{@{\extracolsep{\fill}}lrrrr@{}}
\toprule
\textbf{Planning cost} & \textbf{Agree.} & $\boldsymbol{\rho}$
& \textbf{Flips} & \textbf{SR (\%)} \\
\midrule
Full latent        & 40 & 0.746 & 19.05 & 20 \\
Projected latent   & 90 & 0.975 &  5.29 & 12 \\
Projected + state  & 72 & 0.971 &  5.54 & 10 \\
State only         & 62 & 0.935 &  7.79 & 22 \\
\bottomrule
\end{tabular*}
\vspace{-1em}
\end{table}

Table~\ref{tab:app_stability_utility} shows that Ego projection
raises first-action agreement from 40\% to 90\% and reduces
rank flips from 19.05\% to 5.29\%. Yet the projected cost
reaches the goal in only 6 of 50 episodes, compared with 10
for the Full cost. The state-only cost has lower agreement
than the projected cost but succeeds in 11 episodes.
Agreement measures whether a preference survives an external
change. A planner can consistently prefer an action that makes
little progress toward the goal. This distinction motivates
routing factors into pose, goal, and occupancy estimates, where
their influence on action selection is defined by the task.

\subsection{Planner Implementation}
\label{app:planner_implementation}

The MiniGrid planner converts visual predictions of pose, goal,
and wall occupancy into a graph over grid positions and orientations.
At each step, it propagates the pose belief through the executed
action, mixes in 2\% uniform probability mass, and updates the
belief with the current pose likelihood raised to the power $0.5$.
Wall occupancy is updated as
$\bar W_t=0.8\bar W_{t-1}+0.2W_t$, where $W_t$ is the current
prediction. The goal estimate gives equal weight to the log scores
from the fixed goal image and the mean log scores from observations
seen so far.

From the most probable pose, the planner evaluates left turn,
right turn, and forward actions using Dijkstra costs to the
estimated goal. Each action has unit cost. A forward move into
cell $v$ incurs an additional cost of $12\bar W_t(v)$, and a
candidate successor pose $u'$ incurs a revisit penalty of
$6\max(N_t(u')-1,0)$, where $N_t$ counts visits to estimated
poses. The planner executes one action before updating its
predictions and replanning. All representation variants use
the same belief update, search costs, and 50-step budget.

\begin{figure*}[htbp]
  \centering
  \includegraphics[width=1\linewidth]{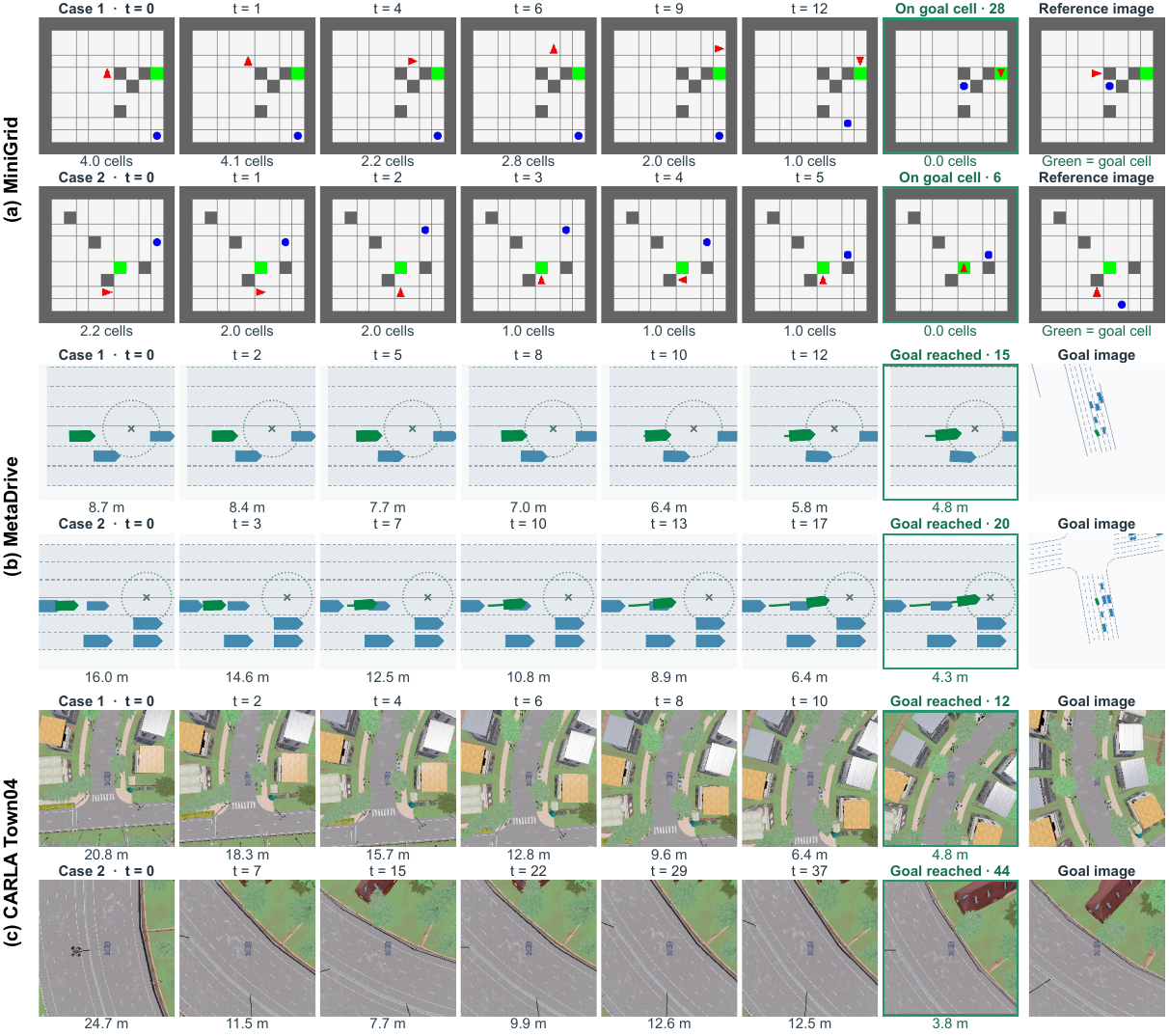}
\caption{\textbf{Successful closed-loop navigation in three environments.}
Each row shows one episode in (a) MiniGrid, (b) MetaDrive, or (c) CARLA.
Labels give the time step and distance to the goal. Green borders mark
the first successful step, and the final column shows the reference image.
Success requires reaching the green goal cell in MiniGrid or entering
the 5\,m goal region in MetaDrive and CARLA.}
\label{fig:navigation_success_cases}
\vspace{-1em}
\end{figure*}

\subsection{Factor Inputs}
\label{app:factor_inputs}

We vary how the factors enter the task readouts while holding
the planner fixed. Full uses the original visual representation.
In the routed variants, Ego corrects the frozen Full pose
prediction, while Scene and Ego, optionally augmented with Exo,
correct the goal and occupancy predictions. Factor-only variants
train new readouts from their designated inputs. Table~%
\ref{tab:app_planner_inputs} includes the additional Full + SE
control, which retains the Full pathway while omitting Exo from
the map adapter.

\begin{table}[htbp]
\centering
\caption{\textbf{Factor inputs to the MiniGrid planner.}
SR is the mean $\pm$ sample SD over five readout training seeds
on the same 150 starts. S, E, and X denote Scene, Ego, and Exo.
The factorizer is fixed across readout seeds within each variant.}
\label{tab:app_planner_inputs}
\small
\begin{tabular*}{\linewidth}{@{\extracolsep{\fill}}llr@{}}
\toprule
\textbf{Readout} & \textbf{Input} & \textbf{SR (\%)} \\
\midrule
Base        & Full             & 78.80 $\pm$ 3.18 \\
Residual    & Full + SE        & 81.20 $\pm$ 2.28 \\
Residual    & Full + SEX       & \textbf{82.40} $\pm$ 1.38 \\
\midrule
Factor only & SEX (basic CCRF) & 81.20 $\pm$ 3.18 \\
Factor only & SEX (final CCRF) & 78.93 $\pm$ 4.15 \\
Factor only & SE  (final CCRF) & 76.80 $\pm$ 5.11 \\
Factor only & Ego (final CCRF) & 75.33 $\pm$ 3.86 \\
\bottomrule
\end{tabular*}
\end{table}

With the final factorizer, SEX alone reaches 78.93\% SR, close
to the Full baseline at 78.80\%. Routing the same factors into
the Full predictions yields the highest mean, 82.40\%.
The paired gain over Full is 3.60 percentage points
(95\% paired $t$ interval across readout seeds,
$[-0.56,\,7.76]$). Adding Exo to the separately trained
Full + SE adapter raises mean SR by 1.20 points. The following
intervention tests whether the trained Full + SEX planner
benefits from receiving the Exo code for the correct episode.

\subsection{Exo Interventions}
\label{app:exo_interventions}

We test whether the trained MiniGrid planner benefits from receiving
the Exo code for the current episode. For each Full + SEX checkpoint,
we evaluate the same 150 starts with the observed Exo code, a code
randomly permuted across active episodes in the same batch, or a zero
vector. Permutation preserves the batch distribution of Exo inputs
while disrupting their episode correspondence. Zeroing removes the
standardized Exo input altogether. In all conditions, the Full pathway
receives the original observation and the adapter weights are fixed.

Table~\ref{tab:app_exo_interventions} shows that the observed and
permuted codes yield the same mean SR of 82.40\%. The paired difference
is 0.00 percentage points (95\% CI $[-1.66,\,1.66]$). Zeroing the code
yields 81.47\% SR, with an observed-minus-zeroed difference of
0.93 points (95\% CI $[-1.14,\,3.01]$). Thus, the higher mean SR of
Full + SEX over Full characterizes the complete routing configuration.
These interventions do not establish an additional average benefit
from matching the explicit Exo code to the episode. The Full pathway
still observes the external object, so the role of external information
in the planner as a whole remains open.

\begin{table}[htbp]
\centering
\caption{\textbf{Inference-time Exo interventions on MiniGrid.}
SR is the mean $\pm$ sample SD over five readout seeds on
150 shared starts. Differences and paired 95\% $t$ intervals
are in percentage points.}
\label{tab:app_exo_interventions}
\small
\begin{tabular*}{\linewidth}{@{\extracolsep{\fill}}lrr@{}}
\toprule
\textbf{Exo input} & \textbf{SR (\%)} &
\textbf{Observed minus variant [95\% CI]} \\
\midrule
Observed & 82.40 $\pm$ 1.38 & -- \\
Zeroed   & 81.47 $\pm$ 1.28 & $+0.93\;[-1.14,\,3.01]$ \\
Permuted & 82.40 $\pm$ 2.14 & $\phantom{+}0.00\;[-1.66,\,1.66]$ \\
\bottomrule
\end{tabular*}
\vspace{-1em}
\end{table}

\begin{figure*}[!htbp]
  \centering
  \includegraphics[width=0.98\linewidth]{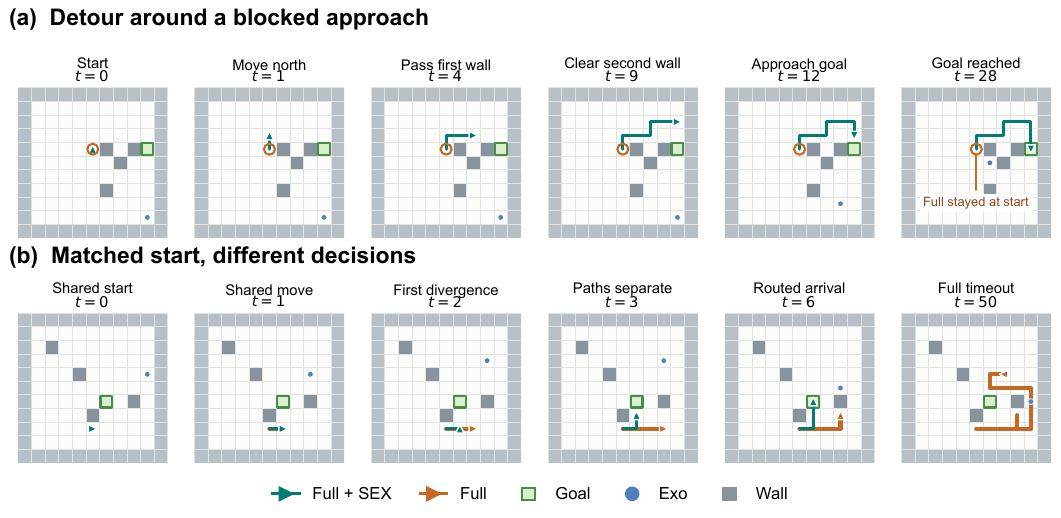}
\caption{\textbf{Closed-loop navigation on MiniGrid.}
(a) Full + SEX initially moves away from the goal to bypass two walls
and reaches it at step 28. Full stays at the start (orange circle)
for all 50 steps.
(b) From the same start and readout seed, the planners first diverge
on the second action. Full + SEX turns left and reaches the goal at
step 6, while Full continues forward and times out at step 50.
Lines show executed paths, and triangles indicate agent headings.}
\label{fig:minigrid_navigation_cases}
\vspace{-1em}
\end{figure*}

\subsection{Qualitative navigation examples.}

Fig.~\ref{fig:navigation_success_cases} shows two successful
episodes in each environment. The distance labels track progress
up to the first step at which the agent reaches the goal cell
in MiniGrid or enters the 5\,m goal region in MetaDrive and CARLA.
The paired MiniGrid examples in Fig.~\ref{fig:minigrid_navigation_cases}
show how individual action choices shape the executed path.
In (a), the start and goal lie on the same
row, with two wall cells between them. The first move by Full + SEX
increases the Euclidean distance to the goal from $4$ to $\sqrt{17}$
cells, yet opens a route around the obstacles. This illustrates why
the planner scores actions using the remaining path cost on its
predicted map: a useful move can temporarily increase direct goal
distance.

In (b), both planners take the same first action. At the next
decision, Full + SEX turns toward the goal while Full advances past
the turn. The routed planner reaches the goal four steps later.
Because the two variants share the same start, readout seed, and
graph search procedure, this case illustrates how a change in the
task readout can alter an early action and the subsequent trajectory.

\section{Task Adaptations}
\label{app:adaptations}

Across tasks, FactorWorld reads the current and goal images through
a frozen visual encoder. The controller determines which predictions
it needs: pose and free space for navigation, joint angles for
feedback control, or object configuration for trajectory search.
State and action labels supervise the corresponding adapters during
training. At execution, the controller uses visual predictions and,
where applicable, previously executed actions.

\begin{figure*}[!htbp]
  \centering
  \includegraphics[width=1\linewidth]{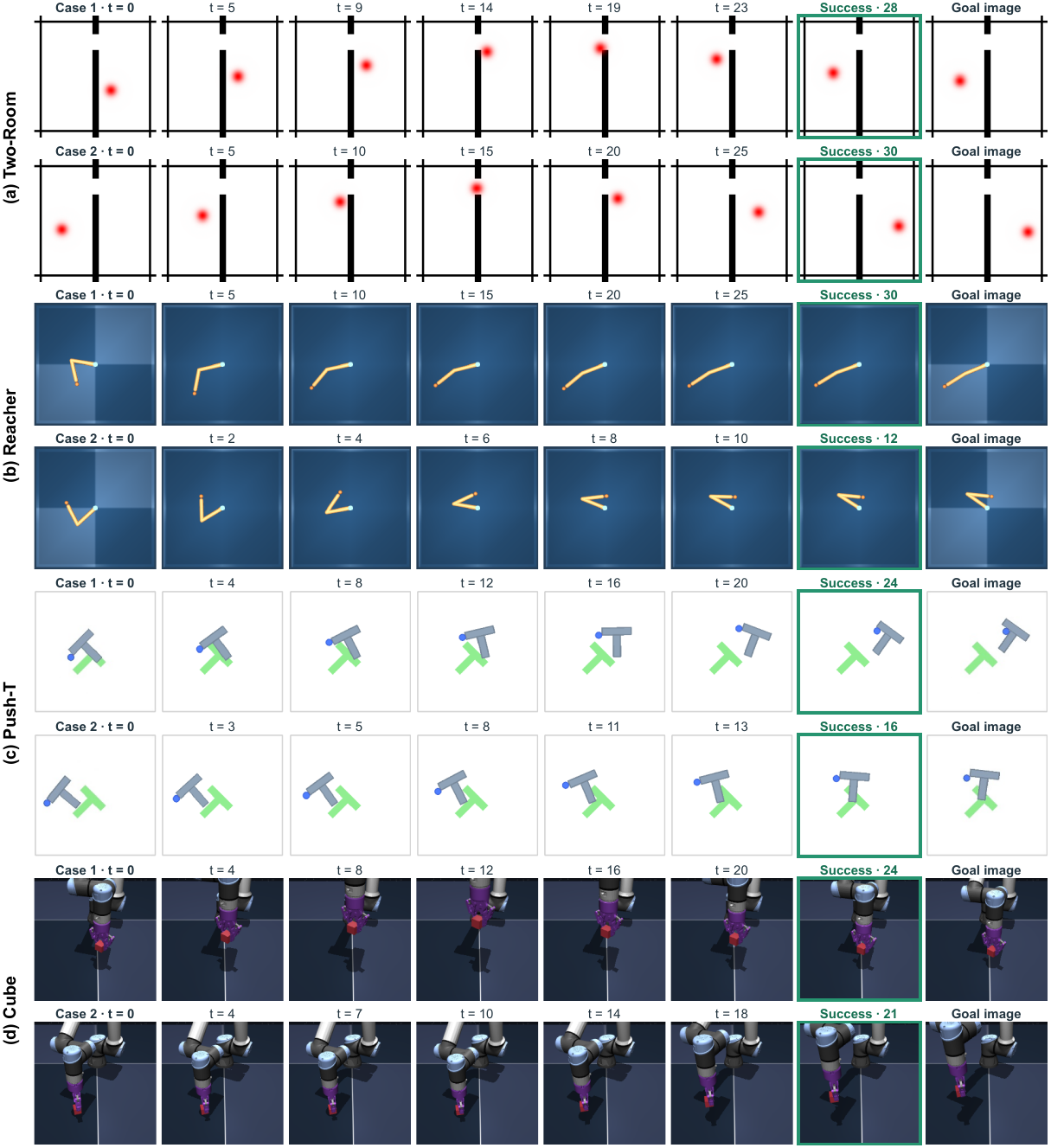}
\caption{\textbf{Closed-loop execution on four continuous control tasks.}
Each row shows one successful confirmation episode in (a) Two-Room,
(b) Reacher, (c) Push-T, or (d) Cube. Frames are labeled by time step.
Green borders mark the first step that satisfies the task's success
criterion. The final column shows the goal image supplied to the controller.}
\label{fig:continuous_control_cases}
\vspace{-0.5em}
\end{figure*}

\subsection{MetaDrive and CARLA}
\label{app:driving_adaptations}

\paragraph{MetaDrive.}
The planner scores predicted trajectories by their progress toward
the goal image and uses a goal-conditioned action prior to guide
CEM. The prior is trained on 25-step action sequences from training
scenes. At each planning step, its prediction is averaged with the
warm start from the previous plan to initialize the candidate
distribution. A prior cost, calibrated using validation action
errors, also guides candidate selection. CEM uses 256 candidates,
12 iterations, and 32 elites over a 25-step horizon. The controller
executes five steps before replanning.

A matched ablation fixes the semantic trajectory cost, action
bounds, feedback interval, and search budget. Adding the prior
raises Full + SEX success from 88.80 $\pm$ 0.30\% to
99.33 $\pm$ 0.00\% across five readout seeds on the fixed
regression set. This 10.53-point gain measures the joint
contribution of prior initialization and prior-based scoring.
Among the 86 starts initially outside the 5\,m goal radius,
the full system reaches the goal in 85 cases per seed.

\paragraph{CARLA.}
The planner uses visual estimates of vehicle pose, goal position,
and local road occupancy to select steering and throttle commands.
A learned motion model propagates the pose belief between
observations. The planner uses the estimated pose to align
successive road predictions, then searches the resulting local
graph for a low-cost path to the goal. This routing gives each
prediction a distinct role: pose aligns the map, occupancy
determines traversal costs, and the goal specifies the search
destination. On the 150 Town04 evaluation starts, Full + SEX
achieves a mean goal-reaching success rate of 98.00\%
(sample SD of 0.67 percentage points across three readout seeds).

\subsection{Two-Room}
\label{app:two_room}

The key planning decision is how to reach the doorway when the
wall blocks a direct path to the goal. A frozen Full + Ego readout
estimates agent and goal positions from their respective images.
The planner extracts obstacles from the current image, expands
them to account for the agent footprint, and uses A* to find a
collision-free path on the predicted map. A lookahead waypoint
provides the next displacement, which is converted to an action
using a scale fitted on training trajectories. The selected
configuration uses an obstacle inflation radius of 8 and a path
lookahead of 3. All three readout seeds complete 50 episodes.

\subsection{Reacher}
\label{app:reacher}

Reacher requires a visual estimate precise enough for joint
feedback. A spatial residual CNN refines the frozen readout's
sine and cosine predictions for each joint. We train the residual
on both recorded images and images re-rendered from the same
training episodes, exposing it to the appearance of the simulator
used at execution. The controller estimates current and target
joint angles from images and converts their angular error into
actions using inverse dynamics fitted to training state--action
pairs. The selected controller uses proportional gain $k_p=250$
and relies on passive damping ($k_d=0$). All three independently
trained residual models complete the same 50 confirmation episodes.

\subsection{Push-T}
\label{app:pusht}

Successful pushing requires the pusher and block to reach their
target positions while the block also reaches its target angle.
We calibrate the factor readout against an image-space estimate
using a residual weight of 0.9 for position and a wrapped residual
for orientation. A learned contact model predicts the effect of
candidate actions on this configuration. The model is trained
on 400,000 contact-balanced one-step transitions, and an action
proposal network is trained on 500,000 trajectory windows.

For predicted step $k$, let $\hat p_k\in\mathbb{R}^4$ contain
the planar positions of the pusher and block, and let
$\hat\theta_k$ denote block orientation. Given the corresponding
goal quantities $p_g$ and $\theta_g$, we score a trajectory by
\begin{equation}
\label{eq:app_pusht_cost}
\begin{aligned}
e_k &= \max\!\left\{
  \frac{\|\hat p_k-p_g\|_2^2}{20^2},
  \frac{\operatorname{wrap}(\hat\theta_k-\theta_g)^2}{(\pi/9)^2}
\right\}, \\
J &= \min_{1\leq k\leq 25} e_k + 0.01\,e_{25}.
\end{aligned}
\end{equation}
The maximum requires both errors to fall within tolerance for
$e_k\leq1$. The minimum credits a trajectory that reaches the
target configuration before its final step. CEM uses 300 candidates,
30 iterations, and 30 elites, replanning after five actions.
With the readout, contact model, and proposal network varied across
training repeats, the system completes 100, 100, and 99 of the
same 100 confirmation episodes.

\subsection{Cube}
\label{app:cube}

In Cube, semantic predictions guide both what the planner scores
and where it begins searching. The visual readout estimates six
joint coordinates and the three-dimensional object position from
each of the current and goal images. Object position defines the
goal cost. The concatenated 18-dimensional state conditions a
proposal network that predicts 25 five-dimensional actions to
initialize CEM. This gives the search access to the robot
configuration while keeping its objective tied to the object.

The proposal network is trained on 175,296 action windows from
the adaptation training episodes. CEM refines its proposal using
300 candidates, 10 iterations, and 30 elites. With the readout
seed fixed, proposal initialization raises success from 74/100
to 91/100 on the same validation episodes. After configuration
selection, the three readout checkpoints share the frozen world
model and proposal network and complete 95, 89, and 91 of the
same 100 confirmation episodes (91.67 $\pm$ 3.06\%).

\subsection{Qualitative Control Examples}

Fig.~\ref{fig:continuous_control_cases} shows how the adapted
controllers reach visual goals across four tasks. In Two-Room,
the two agents approach the doorway from opposite sides. Both
first move toward the opening, cross the wall, and then turn
toward the goal, reaching it at steps 28 and 30. These trajectories
illustrate how the obstacle map shapes the route even when the
goal position is known from the visual readout. In Reacher, the
relevant quantity is the arm configuration. In the first example,
the arm changes from a bent pose to a nearly extended one over
30 steps. The second reaches a bent target pose in 12 steps.

The manipulation tasks show why intermediate frames are useful
for interpreting success. In Push-T, the blue pusher approaches
the gray block from the left in both episodes. As the pusher
advances, the block translates and rotates toward the pose shown
in the goal image. The two episodes reach their targets at steps
24 and 16. In Cube, the gripper and cube move through intermediate
configurations before the environment registers success at steps
24 and 21. Across these tasks, the same visual goal format guides
different decisions, from choosing a traversable route to selecting
actions that change an object's pose through contact.

\section{Limitations}
\label{app:limitations}

FactorWorld uses counterfactual pairs to ask how a representation
changes when one trajectory varies while the other is held fixed.
Planning in a reactive environment introduces a different question:
how would an external actor respond if the agent chose a different
action? At a merge, for example, braking by another vehicle may
carry little information about the agent's current pose yet determine
whether moving forward is feasible. Stability under changes of the
external trajectory leaves this action-dependent response unresolved.
Extending FactorWorld to such interactions would require predicting
how external behavior changes under candidate actions and testing
those predictions against subsequent observations.



The semantic interface also limits which distinctions reach the
planner. Two scenes can share the same ego pose, goal, and current
occupancy yet require different actions if another vehicle is
approaching the crossing in one scene and leaving it in the other.
An open question is how an agent can detect that its current
interface is insufficient. One direction is to use discrepancies
between predicted and observed action outcomes to identify an
omitted quantity, such as time to occupancy, and learn when the
task requires it.